\documentclass[runningheads]{llncs}

\usepackage{eccv}

\usepackage{eccvabbrv}

\usepackage{graphicx}
\usepackage{booktabs}

\usepackage[accsupp]{axessibility} 

\usepackage{multirow}

\usepackage{algorithm} 
\usepackage{algpseudocode}
\usepackage{amsmath} 

\usepackage{wrapfig}
\usepackage[table]{xcolor}
\usepackage{tabularx}

\usepackage{pifont} 
\newcommand{\cmark}{\ding{51}}
\newcommand{\xmark}{\ding{55}}

\usepackage{hyperref}

\usepackage{orcidlink}

\begin{document}

\title{CUST : Clustered Unit-level Similarity Transformer for Lightweight Image Super-Resolution} 

\titlerunning{CUST for Lightweight Image Super-Resolution}

\author{Jeongsoo Kim\inst{1} \orcidlink{0009-0009-9864-9601}}


\authorrunning{J.~Kim}


\institute{
Independent Researcher \\
\email{jeongskim512@gmail.com}}

\maketitle



\begin{abstract}
  Recently, Vision Transformer (ViT)-based models have exhibited remarkable performance in image super-resolution. However, the quadratic computational complexity of ViTs with respect to spatial resolution severely constrains their efficiency, leading to high latency and massive memory consumption. To alleviate this, various window-based attention mechanisms have been proposed; yet, they inherently compromise the long-range dependency modeling that is the primary advantage of ViTs. To overcome these limitations, we propose the Clustered Unit-level Similarity Transformer (CUST), a novel architecture that efficiently integrates global and local information. Specifically, CUST enables each patch to aggregate and attend to similar patches within a broadened regional scope outside its local window, thereby capturing extensive contextual understanding. Furthermore, it employs overlapping attention windows to capture local dependencies, while explicitly extracting high-frequency details by computing the residual difference between the original features and their downsampled-upsampled counterparts. Comprehensive experiments demonstrate that our proposed model achieves a practical balance between computational efficiency and restoration performance. It achieves a lower memory footprint and faster inference speed compared to recent global context or lightweight models under realistic constraints. Code is available at \url{https://github.com/jwgdmkj/CUST}.
  
\end{abstract}

\section{Introduction}
\label{intro}
\begin{figure*}[t]
    \centering
    \includegraphics[width=0.65\textwidth]{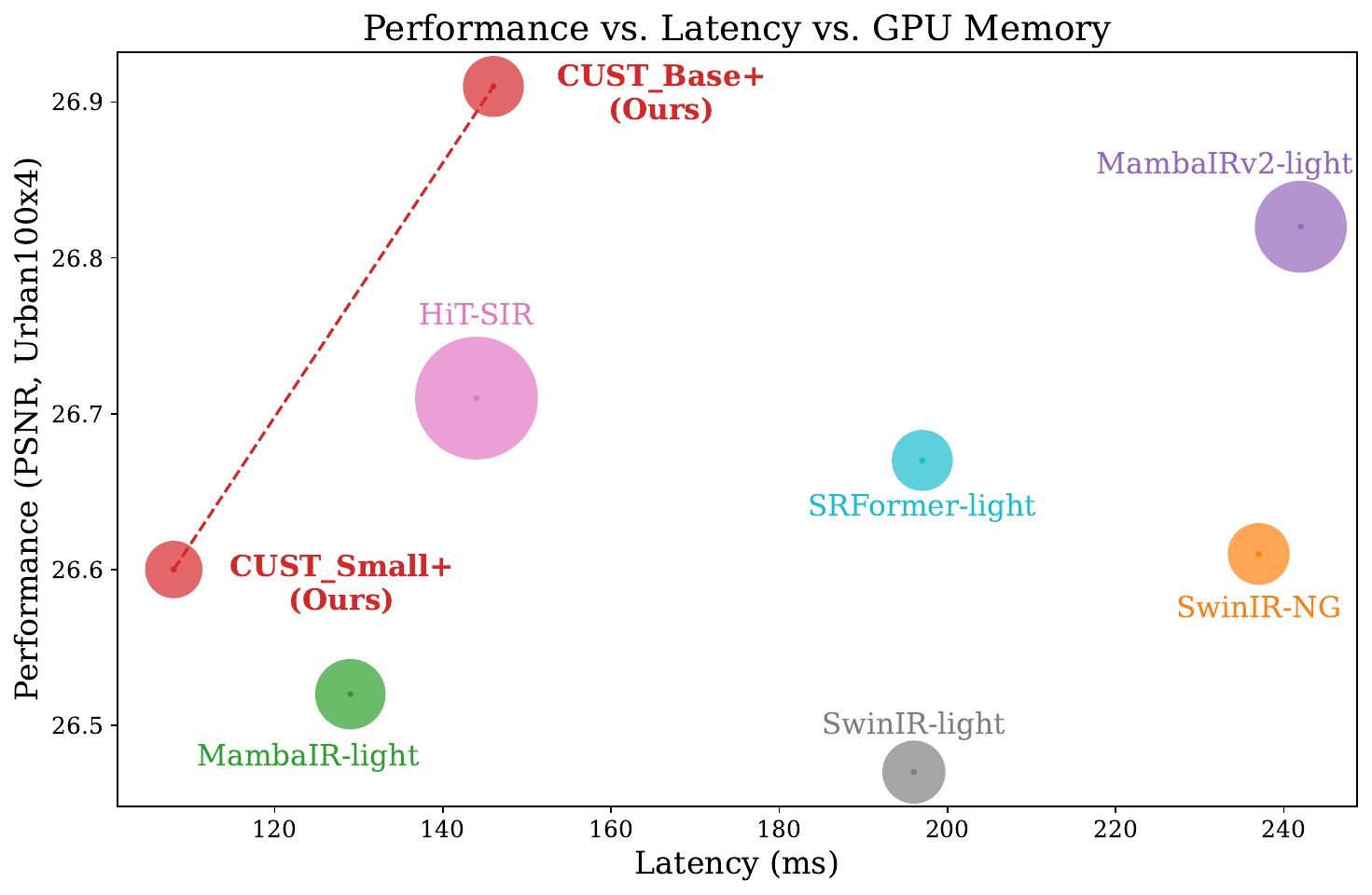} 

    \caption{
    Efficiency-Performance trade-off analysis on Urban100 ($\times$4).
    }
    \label{fig:teaser}
\end{figure*}

Single Image Super-Resolution (SISR), which aims to reconstruct a HR (High-Resolution) image from a given LR (Low-Resolution) counterpart, is a fundamental and actively researched area in computer vision. SISR has been widely applied in various real-world scenarios, ranging from medical imaging requiring precise diagnosis and digital photography for restoring old photos to high-magnification digital zoom in smartphones~\cite{phonesr, medicalsr}.

Since the pioneering work of SRCNN~\cite{srcnn}, which first introduced Convolutional Neural Networks (CNNs) to SISR, numerous CNN-based models have been proposed~\cite{srgan, san, csfm, RCAN}. However, CNNs suffer from a fundamental limitation regarding their restricted receptive fields, making it difficult to capture long-range dependencies effectively. Although various approaches have attempted to overcome this issue by stacking deeper layers~\cite{rdn, edsr, vdsr}, this strategy often leads to increased computational costs and model complexity~\cite{safmn}.

Following the remarkable success of the attention mechanism in Natural Language Processing (NLP)~\cite{Attn}, various models adopting this paradigm have emerged in the computer vision domain~\cite{ViT, swin, cswin, pvt, poolformer}. The attention mechanism effectively captures long-range dependencies, addressing the shortcomings of CNNs. However, Vision Transformers (ViTs) suffer from quadratic computational complexity relative to input resolution~\cite{orthogonal, ngswin, qvit, cat}. To alleviate this, Window-based Self-Attention (WSA) divides images into fixed-size windows~\cite{srformer, swinir}; however, this restricts the receptive field and undermines the capture of long-range dependencies~\cite{grl, hat, fastervit, lmlt}. Subsequent attempts to mitigate these limitations—such as generating super-tokens~\cite{spin, catanet} or employing large-window lightweighting strategies~\cite{ipg, pft, atd}—introduce new bottlenecks. Super-token methods are often hindered by slow iterative generation or memory-intensive similarity calculations, while many lightweight models prioritize parameter or FLOPs reduction over practical hardware metrics like inference latency and peak memory usage. Consequently, achieving a practical resource-aware Image Super-Resolution (SR) suitable for resource-constrained deployment remains a significant challenge.



To address these challenges, we propose the Clustered Unit-level Similarity Transformer (CUST). Our CUST is composed of the Cross-window Affinity Neighbor Attention (CANA) module, which aggregates and operates on semantically similar patches beyond window boundaries, and the Multi-frequency Error-driven Dense Attention (MEDA) module, which restores high-frequency details and reinforces local information through multi-scale error signals. Specifically, CANA aggregates semantically similar patches across window boundaries by sorting and clustering them based on their affinity to pooled window tokens. This mechanism enables the model to effectively capture long-range context beyond fixed constraints.


While CANA provides a long-range context, the subsequent MEDA module restores intricate local fidelity by extracting high-frequency error signals from multi-scale feature differences. These errors, refined via dilated convolutions, explicitly guide overlap window attention toward salient structures like edges while simultaneously expanding the receptive field. By capturing diverse information beyond restricted window areas, CUST achieves superior restoration performance with a lower memory footprint than super-token models and enhanced hardware efficiency compared to conventional WSA-based designs. This dual emphasis on performance and hardware efficiency ensures practical utility for real-world hardware deployment, as illustrated in Fig.~\ref{fig:teaser}.

Our extensive experiments demonstrate that the proposed CUST architecture achieves a favorable balance between model complexity and restoration accuracy. Compared to other state-of-the-art window-based attention (WSA) models~\cite{swinir, srformer}, CUST-Base captures a significantly broader spatial context while reducing average inference latency by 26.7\%. Furthermore, in comparison to CATANet~\cite{catanet}, our model achieves an average of 83\% memory reduction across all scales while maintaining comparable inference speeds. Notably, it delivers an average PSNR improvement of 0.094 dB across all benchmark datasets at $\times$4 scale, demonstrating practical efficiency and restoration fidelity simultaneously.
\section{Related Works}

\textbf{Single Image Super-Resolution} (SISR) is a fundamental research problem in the field of computer vision. With the advancement of deep learning, diverse architectures have been introduced. Since SRCNN~\cite{srcnn} pioneered end-to-end high-resolution reconstruction using a three-layer CNN, subsequent models such as IRCNN~\cite{ircnn} and VDSR~\cite{vdsr} have emerged, significantly outperforming traditional interpolation methods. However, CNN-based models inherently suffer from restricted receptive fields, limiting their ability to exploit global context. To address this, attention mechanisms were adopted for SISR. IPT~\cite{IPT} demonstrated superiority over CNNs. However, standard self-attention suffers from quadratic complexity relative to input resolution. SwinIR~\cite{swinir} mitigated this through Window-based Self-Attention (WSA), yet restricted the receptive field to local windows, failing to fully exploit global context. To overcome these local constraints, we propose a novel architecture that groups windows into broader search regions and employs a similarity-based cross-window clustering mechanism to capture long-range dependencies efficiently.


\noindent\textbf{Efficient Image Super-Resolution} aims to improve the computational efficiency of high-performance SR models. Although deep learning-based SISR has achieved remarkable performance advancements, the accompanying surge in computational cost hinders its deployment on resource-constrained hardware environments. FSRCNN~\cite{fsrcnn} pioneered this effort by replacing the bicubic interpolation in SRCNN~\cite{srcnn} with a deconvolution layer, significantly accelerating inference speed. ELAN~\cite{elan} utilized group-wise multi-scale attention. SPIN~\cite{spin} adopted superpixel-based interactions. CATANet~\cite{catanet} efficiently captured broad-range attention by generating tokens that represent global image regions. While many state-of-the-art models offer lightweight variants via hyperparameter tuning~\cite{srformer, swinir, ipg}, they often prioritize parameter reduction over practical metrics such as inference latency and peak GPU memory footprint. Moreover, existing token-generation methods frequently encounter efficiency bottlenecks due to iterative computations. In contrast, we demonstrate that refining high-frequency details through multi-scale error signals allows for a reduction in window size and memory footprint while maintaining or even enhancing restoration performance.

\section{Method}
\label{method}
\begin{figure*}[t]
    \centering
    \begin{minipage}[c]{0.65\textwidth}
        \centering
        \includegraphics[width=\linewidth]{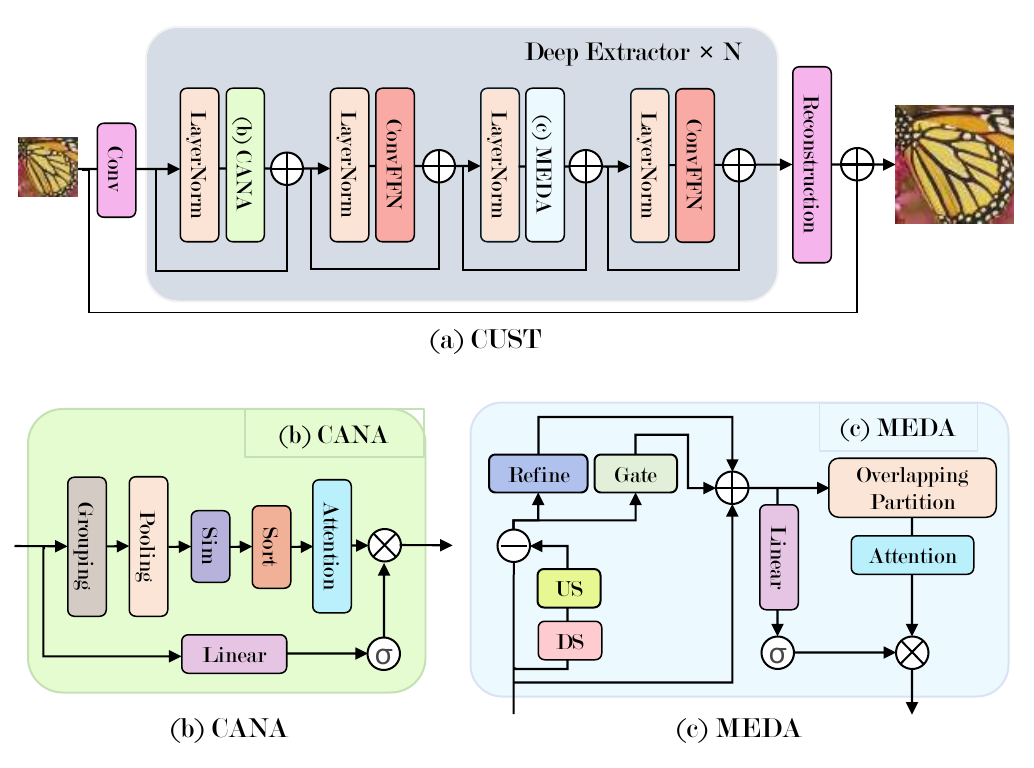}
    \end{minipage}
    \hfill 
    \begin{minipage}[c]{0.28\textwidth}
        \caption{
            \textbf{The architecture of CUST.} 
            (a) The overall framework of CUST. 
            (b) The structure of Cross-window Affinity Neighbor Attention (CANA) module. 
            (c) The structure of Multi-frequency Error-driven Dense Attention (MEDA) module.
        }
        \label{fig:arch}
    
    \end{minipage}
    \vspace{-3mm}
\end{figure*}

\subsection{Overall Architecture}

Figure~\ref{fig:arch}(a) illustrates the overall architecture of our proposed \textbf{Clustered Unit-level Similarity Transformer (CUST)}. Given an input low-resolution (LR) image $I_{LR} \in \mathbb{R}^{H \times W \times 3}$, a $3\times3$ convolutional layer first extracts shallow features $F_0 \in \mathbb{R}^{H \times W \times C}$. Subsequently, these features are passed through a series of deep feature extractor blocks to extract deep-level features. Each deep feature extractor block comprises a Cross-window Affinity Neighbor Attention (CANA) module, a Multi-frequency Error-driven Dense Attention (MEDA) module, two Convolutional Feed-Forward Network (ConvFFN) modules, and four layer normalization ~\cite{ln} layers. Specifically, the CANA module performs a similarity-based clustering operation to interact with semantically related patches beyond window boundaries, capturing long-range dependencies. In contrast, the MEDA module focuses on reinforcing local correlations and restoring high-frequency details using multi-scale error signals. Finally, the deep features are transformed into a high-quality image through a reconstruction module~\cite{espcn}, and the final high-resolution output is obtained by adding the global residual of the input $I_{LR}$. Detailed mechanisms of the CANA and MEDA modules are elaborated in Section~\ref{cana} and Section~\ref{meda}, respectively.


\subsection{CANA : Cross-window Affinity Neighbor Attention}
\label{cana}
\begin{figure*}[t]
    \centering
    \includegraphics[width=0.9\textwidth]{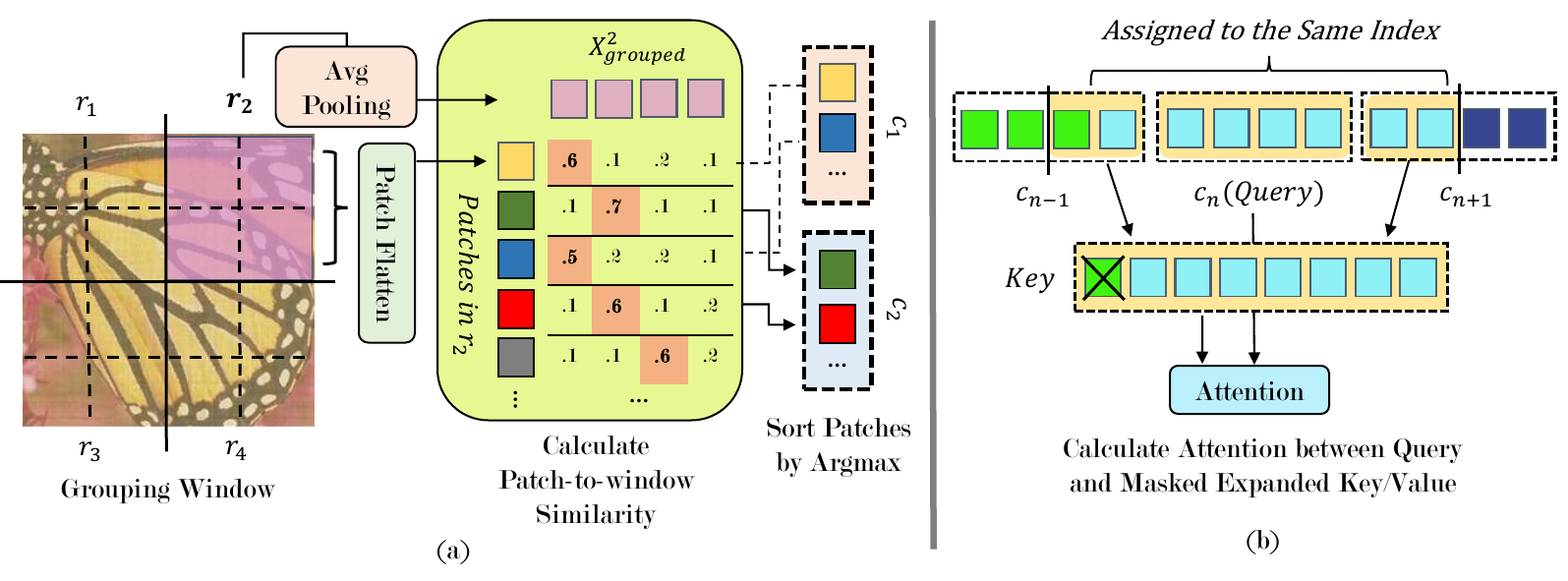} 

    \caption{
    Illustration of the proposed Cross-window Affinity Neighbor Attention (CANA) mechanism. (a) Patches within a search region are clustered based on their similarity to pooled window tokens. (b) To prevent information disconnection at cluster boundaries, Key and Value sets are expanded by incorporating patches from adjacent clusters that share the same window token affinity.
    }
    \label{fig:mechanism}
    \vspace{-3mm}
\end{figure*}

Our proposed Cross-window Affinity Neighbor Attention (CANA) is illustrated in detail in Fig.~\ref{fig:arch}(b) and Fig.~\ref{fig:mechanism}(a).

\paragraph{Representative Token Generation.} First, given a feature map $I \in \mathbb{R}^{H \times W \times C}$, we partition it into multiple search regions $\mathcal{R} = \{r_1, r_2, \dots, r_M\}$ with a size of $GS \times GS$, where the grid size $GS$ is a multiple of the window size $ws$. Within each search region  $r_m \in \mathbb{R}^{GS \times GS \times C}$, we generate Representative Window Tokens $X^{m}_{grouped} \in \mathbb{R}^{N \times C}$ to summarize the local context of each window. Here, we apply a non-overlapping average pooling operation with a kernel size and stride of $ws$, to downsample each $ws \times ws$ window into a $ 1 \times 1$ token. This formulation is expressed as Eq.~\ref{eq:xgroup}:

\begin{equation}
\label{eq:xgroup}
    {X}^{m}_{grouped} = \operatorname{AvgPool}_{ws}(r_m) 
    \in \mathbb{R}^{N \times C}
\end{equation}

where $N = (GS/ws)^2$ denotes the number of partitioned windows (and thus the number of pooled window tokens) within the search region.

\paragraph{Similarity-based Clustering.} To establish a robust semantic grouping mechanism, we adapt the similarity-based patch clustering paradigm of CATANet~\cite{catanet} into our window-group-level framework. Specifically, we compute the dot-product similarity between all individual spatial patches within $r_m$ and the representative window tokens $X^{m}_{grouped}$. Through an $\operatorname{Argmax}$ operation along the token dimension, each patch is assigned to the index of the window representative that yields the highest similarity. By gathering patches that share the same index through an $\operatorname{Argsort}$ operation, a cluster $\mathcal{C}_n$ sharing the identical index is formed. Through this, similar patches scattered globally become adjacent to each other. Because the total number of mutually related patches across the search region can often exceed the predefined capacity of a single cluster, some highly relevant features are partitioned into neighboring clusters. To prevent such information disconnection at the cluster boundaries and to expand the receptive field, we extend the Key and Value regions, as illustrated in Fig.~\ref{fig:mechanism}(b). Specifically, when using the current cluster $\mathcal{C}_n$ as the Query, we fetch the nearest halves of the patches from the adjacent clusters $\mathcal{C}_{n-1}$ and $\mathcal{C}_{n+1}$ to expand the Key and Value sets. Since the entire sequence is sorted by semantic affinity, the index-adjacent clusters $\mathcal{C}_{n-1}$ and $\mathcal{C}_{n+1}$ represent the closest semantic matches rather than mere spatial neighbors. Let $\mathcal{C}^{rear}$ and $\mathcal{C}^{front}$ denote the second and first half of a cluster, respectively. The expanded set is formulated as Eq.~\ref{eq:cana_chunk}:

\begin{equation}
\label{eq:cana_chunk}
    \mathbf{K}_{exp}, \mathbf{V}_{exp} = \operatorname{Concat}(\mathcal{C}_{n-1}^{rear}, 
                                        \mathcal{C}_n, 
                                        \mathcal{C}_{n+1}^{front})
\end{equation}


\paragraph{Attention with Masking.} To exclude interactions with semantically unrelated patches among the merged candidates, we apply an attention masking strategy. The attention mask $\mathbf{M}$ is defined such that it takes a value of 0 when a key token $j$ shares the same window ID as the query token $i$—specifically, when both tokens identify the same window as their most similar neighbor—and $-\infty$ otherwise. This effectively restricts references between patches that do not share the same window ID. Finally, the features are adaptively modulated through a learned gate $\mathbf{G}_n$~\cite{gateattn}, and the overall operation is formulated as follows:

\begin{equation}
\label{eq:cana_attn}
    \mathbf{Y}_n = \operatorname{GELU}\left( \operatorname{Softmax}\left(\frac{\mathbf{Q}_n \mathbf{K}_{exp}^T}{\sqrt{d}} + \mathbf{M} \right)\mathbf{V}_{exp} \right) \odot \mathbf{G}_n
\end{equation}

This process makes each patch overcomes the constraints of a fixed local window and interacts with semantically similar patches, thereby effectively capturing long-range dependencies.

\subsection{MEDA : Multi-frequency Error-driven Dense Attention}
\label{meda}
\begin{algorithm}[t]
\caption{Multi-frequency Error-driven Dense Attention (MEDA)}
\label{algo:meda}
\begin{algorithmic}[1]
    \Require Input feature $\mathbf{X} \in \mathbb{R}^{H \times W \times C}$ 
    \Ensure Refined feature $\mathbf{Y}$

    \Statex \textbf{Step 1: Frequency-aware Error Extraction}
    \State $\mathbf{X}_{low} \leftarrow \operatorname{Up}(\operatorname{Down}(\mathbf{X}))$ \Comment{Down-sampling and Up-sampling}
    \State $\mathbf{X}_{error} \leftarrow \mathbf{X} - \mathbf{X}_{low}$ \Comment{Extract High-frequency Residuals}

    \Statex \textbf{Step 2: Error Refinement \& Gating}
    \State $\mathbf{F}_{refine} \leftarrow \operatorname{Conv}_{1\times1}(\operatorname{GELU}(\operatorname{DilConv}_{3\times3}(\mathbf{X}_{error})))$ \Comment{Dilated Context Aggregation}
    \State $\mathbf{G} \leftarrow \sigma(\operatorname{Conv}_{1\times1}(\mathbf{X}_{error}))$ \Comment{Generate Spatial Gating Map}

    \Statex \textbf{Step 3: Feature Compensation}
    \State $\mathbf{X}_{refined} \leftarrow \mathbf{X} + \alpha \cdot (\mathbf{F}_{refine} \odot \mathbf{G})$ \Comment{$\odot$: Element-wise Product}

    \Statex \textbf{Step 4: Local Detail Refinement}
    \State $\mathbf{Y} \leftarrow \operatorname{OverlapSelfAttention}(\mathbf{X}_{refined})$

    \State \Return $\mathbf{Y}$
\end{algorithmic}
\end{algorithm}

While CANA captures global long-range dependencies by overcoming window constraints, the subsequent MEDA module refines intricate local details within the window. The detailed architecture and algorithm of the module are illustrated in Fig.~\ref{fig:arch}(c) and Algorithm~\ref{algo:meda}, respectively. Here, the Overlap Self Attention in Algorithm~\ref{algo:meda} is adopted from HPI-Net~\cite{hpinet}.


\paragraph{Motivation.}

Generally, in window-based attention mechanisms, reducing the window size restricts the receptive field, which leads to performance degradation. To mitigate this, various models have employed overlapping windows to facilitate information exchange across boundaries~\cite{catanet, spin, hpinet}. However, these methods suffer from a sharp increase in computational overhead as the window size and overlap area expand. To address this, we adopt overlapping window attention while proposing an efficient information correction method that leverages high-frequency refinement based on multi-scale errors, even within limited window regions. The proposed MEDA is formulated as shown in Eq.~\ref{eq:meda}.

\begin{equation}
\label{eq:meda}
    \mathbf{X_{refined}} = \mathbf{X} + \alpha \cdot 
    (\operatorname{Refiner}(\mathbf{X}_{error}) \odot 
    \operatorname{Gate}(\mathbf{X}_{error}))
\end{equation}

\paragraph{Algorithm.}

The MEDA module is computed through the following procedure. First, given the input feature $\mathbf{X}$ from CANA, the error feature $\mathbf{X}_{error}$ is generated by calculating the residual between $\mathbf{X}$ and its downsampled-then-upsampled version: $\mathbf{X}_{error} = \mathbf{X} - \operatorname{Up}(\operatorname{Down}(\mathbf{X}))$. This error feature captures high-frequency information, serving as an explicit guide that highlights essential edge and texture locations for reconstruction. Next, $\mathbf{X}_{error}$ is processed through the \textbf{Error Refiner} and the \textbf{Spatial Gate}. The \textbf{Error Refiner}, composed of dilated convolutions, transforms simple pixel-wise differences into semantically rich contextual features. By enlarging the receptive field with minimal computational overhead, it generates sophisticated error features that account for neighboring texture flows. Simultaneously, the \textbf{Spatial Gate}, implemented with $1 \times 1$ convolutions, determines the specific locations and intensities where corrections are required. The resulting refined error is multiplied by a learnable scaler $\alpha$ and added to the original feature $\mathbf{X}$, providing guidance on which regions require focused reconstruction. Finally, overlap window-based self-attention is applied to further refine local details and enhance image quality.
\section{Experiments}

\subsection{Settings}

\begin{table*}[t!]
	\caption{
    Quantitative comparison of CUST-Base and CUST-Base+ with state-of-the-art lightweight SISR models. The best and second-best performances are highlighted in \textcolor{red}{red} and \textcolor{blue}{blue}, respectively.}
\vspace{-3mm}
	\label{tab:base}
	\renewcommand\arraystretch{1.1}
	\begin{center}
		\resizebox{\textwidth}{!}{
			\begin{tabular}{| c | c | c | c | c | c | c | c | c |}
				\hline
				Scale & Method & \#Params & \#FLOPs & Set5 & Set14 & B100 & Urban100 & Manga109 \\
				\hline

                \multirow{14}*{$\times 2$} 
                & SwinIR-light~\cite{swinir}     &910K & 244G 
                &38.14/0.9611
                &33.86/0.9206
                &32.31/0.9012
                &32.76/0.9340
                &39.12/0.9783 \\


                ~ & ELAN-light~\cite{elan} &621K &203G 
                &38.17/0.9611
                &33.94/0.9207
                &32.30/0.9012 
                & 32.76/0.9340
                &39.11/0.9782 \\

                ~ & HPI-Net~\cite{hpinet} & 783K & 429G 
               & 38.12/0.9605 
               & 33.94/0.9209 
               & 32.31/0.9013 
               & 32.85/0.9346 
               & 39.08/0.9771 \\
                
                ~ & SwinIR-NG~\cite{ngswin} & 1,181K & 274G 
                & 38.17/0.9612 
                & 33.94/0.9205 
                & 32.31/0.9013 
                & 32.78/0.9340
                & 39.20/0.9781 \\
                
                ~ &SRformer-Light~\cite{srformer}       & 853K & 236G 
                & 38.23/0.9613
                &33.94/0.9209
                &32.36/0.9019
                &32.91/0.9353
                &39.28/0.9785 \\ 

               ~ & SPIN~\cite{spin} &  497K & 114G
               & 38.20/\textcolor{blue}{0.9615}
               & 33.90/0.9215 
               & 32.31/0.9015
               & 32.79/0.9340
               & 39.18/0.9784 \\

                ~ & OSFFNet~\cite{osffnet} & 516K & 83G 
                & 38.11/0.9610
                & 33.72/0.9190 
                & 32.29/0.9012 
                & 32.67/0.9331 
                & 39.09/0.9780 \\

                ~ & HIT-SIR~\cite{hit} & 772K & 210G
                & 38.22/0.9613 
                & 33.91/0.9213 
                & 32.35/0.9019 
                & 33.02/0.9365 
                & 39.38/0.9782 \\

                ~ & SMFANet+~\cite{smfanet} & 480K & 108G 
                & 38.19/0.9611 
                & 33.92/0.9207 
                & 32.32/0.9015 
                & 32.70/0.9331 
                & 39.46/\textcolor{red}{0.9787} \\
                

                ~ & MambaIR-light~\cite{mambair} & 847K & 227G 
                &  38.13/0.9610 
                & 33.95/0.9208 
                & 32.31/0.9013 
                & 32.85/0.9349 
                & 39.20/0.9782 \\

                ~ & MambaIRv2-light~\cite{mambairv2} & 774K & 286G
                & 38.26/\textcolor{blue}{0.9615}
                & \textcolor{blue}{34.09/0.9221} 
                & 32.36/0.9019 
                & \textcolor{red}{33.26}/\textcolor{red}{0.9378}
                & 39.35/0.9785 \\

                ~ & CATANet~\cite{catanet} & 477K & 135G
                & 38.28/\textcolor{red}{0.9617}    
                & 33.99/0.9217
                & \textcolor{blue}{32.37}/0.9023
                & 33.09/0.9372
                &  39.37/0.9784 \\

                \cline{2-9}
                ~ & \textbf{CUST-Base(Ours)} & 682K & 291G
                & \textcolor{blue}{38.31}/\textcolor{red}{0.9617}
                & 34.08/\textcolor{blue}{0.9221} 
                &  \textcolor{red}{32.40/0.9026}
                &  \textcolor{blue}{33.25}/0.9375
                & \textcolor{blue}{39.47}/\textcolor{blue}{0.9785}\\

                ~ & \textbf{CUST-Base+(Ours)} & 682K & 314G
                & \textcolor{red}{38.32/0.9617} 
                & \textcolor{red}{34.10/0.9228}
                & \textcolor{red}{32.40}/\textcolor{blue}{0.9024}
                & 33.21/\textcolor{blue}{0.9377}
                &  \textcolor{red}{39.49}/\textcolor{blue}{0.9785}\\
                
				\hline
				\hline


                \multirow{14}*{$\times 3$} 
                & SwinIR-light~\cite{swinir}   &918K & 111G 
                &34.62/0.9289
                &30.54/0.8463
                &29.20/0.8082
                &28.66/0.8624
                &33.98/0.9478 \\

                
                ~ & ELAN-light~\cite{elan} &629K &90G 
                &34.61/0.9288
                &30.55/0.8463
                &29.21/0.8081
                &28.69/0.8624
                &34.00/0.9478 \\ 

                ~ & HPI-Net~\cite{hpinet} & 924K & 277G
               &  34.70/0.9289 
               & 30.63/0.8480 
               & 29.26/0.8104 
               & 28.93/0.8675 
               & 34.31/0.9487 \\

                ~ & SwinIR-NG~\cite{ngswin} & 1,190K & 114G
                & 34.64/0.9293 
                & 30.58/0.8471
                & 29.24/0.8090 
                & 28.75/0.8639
                & 34.22/0.9488 \\
                
                ~ &SRformer-Light~\cite{srformer}  & 861K & 105G
                & 34.67/0.9296
                &30.57/0.8469
                &29.26/0.8099
                &28.81/0.8655
                &34.19/0.9489 \\

                ~ & SPIN~\cite{spin} & 569K & 58G
                & 34.65/0.9293 
                & 30.57/0.8464 
                & 29.23/0.8089 
                & 28.71/0.8627
                & 34.24/0.9489 \\

                ~ & OSFFNet~\cite{osffnet} & 524K & 38G 
                & 34.58/0.9287 
                & 30.48/0.8450 
                & 29.21/0.8080 
                & 28.49/0.8595 
                & 34.00/0.9472 \\

                ~ & HIT-SIR~\cite{hit} & 780K & 94G
                &  34.72/0.9298 
                & 30.62/0.8474 
                & 29.27/0.8101 
                & 28.93/0.8673 
                & 34.40/0.9496 \\

                ~ & SMFANet+~\cite{smfanet} & 487K & 48G 
                & 34.66/0.9292 
                & 30.57/0.8461 
                & 29.25/0.8090 
                & 28.67/0.8611 
                & 34.45/0.9490 \\
                

                ~ & MambaIR-light~\cite{mambair} & 913K & 149G
                & 34.63/0.9288 
                & 30.54/0.8459 
                & 29.23/0.8084 
                & 28.70/0.8631 
                & 34.12/0.9479 \\

                ~ & MambaIRv2-light~\cite{mambairv2} & 781K & 127G
                &  34.71/0.9298 
                & 30.68/0.8483
                & 29.26/0.8098 
                & 29.01/0.8689
                & 34.41/0.9497 \\
                
                ~ & CATANet~\cite{catanet} & 550K & 60G
                & 34.75/\textcolor{blue}{0.9300}   
                & 30.67/0.8481
                & 29.28/0.8101
                & 29.04/0.8689
                & 34.40/\textcolor{blue}{0.9499} \\

                \cline{2-9}
                ~ & \textbf{CUST-Base(Ours)} & 754K & 138G
                & \textcolor{blue}{34.76}/\textcolor{red}{0.9303} 
                & \textcolor{red}{30.73/0.8488}
                & \textcolor{red}{29.34/0.8116}
                & \textcolor{red}{29.11/0.8699}
                & \textcolor{blue}{34.54}/0.9497 \\

                ~ & \textbf{CUST-Base+(Ours)} & 754K & 147G
                & \textcolor{red}{34.79/0.9303}
                & \textcolor{blue}{30.71/0.8485}
                & \textcolor{blue}{29.33/0.8114}
                & \textcolor{blue}{29.10}/\textcolor{blue}{0.8698}
                & \textcolor{red}{34.57}/\textcolor{red}{0.9500}\\
                
    		      \hline
				\hline


                \multirow{14}*{$\times 4$} 
				& SwinIR-light~\cite{swinir}  &930K & 64G
                &32.44/0.8976
                &28.77/0.7858
                &27.69/0.7406
                &26.47/0.7980
                &30.92/0.9151 \\

    
				~ & ELAN-light~\cite{elan} &640K &54G 
                &32.43/0.8975
                &28.78/0.7858
                &27.69/0.7406
                &26.54/0.7982
                &30.92/0.9150 \\ 

                ~ & HPINet~\cite{hpinet} & 896K & 137G
                & 32.60/0.8986 
                & 28.87/0.7874 
                & 27.73/0.7419 
                & 26.71/0.8043 
                & 31.19/0.9161 \\

                ~ & SwinIR-NG~\cite{ngswin} & 1,201K & 63G
                & 32.44/0.8980 
                & \textcolor{black}{28.83}/0.7870 
                & 27.73/0.7418 
                & 26.61/0.8010 
                & 31.09/0.9161 \\
    
                ~ &SRformer-Light~\cite{srformer}  & 873K & 63G 
                & 32.51/0.8988
                & 28.82/\textcolor{black}{0.7872} 
                & 27.73/0.7422
                & 26.67/0.8032
                & 31.17/0.9165\\

                ~ & SPIN~\cite{spin} & 555K & 42G 
                & 32.48/0.8983
                & 28.80/0.7862
                & 27.70/0.7415
                & 26.55/0.7998 
                & 30.98/0.9156 \\

                ~ & OSFFNet~\cite{osffnet} & 537K & 22G
                & 32.39/0.8976 
                & 28.75/0.7852 
                & 27.66/0.7393
                & 26.36/0.7950 
                & 30.84/0.9125 \\
    
                ~ & HIT-SIR~\cite{hit} & 792K & 54G
                & 32.51/0.8991 
                & 28.84/0.7873 
                & 27.73/0.7424 
                & 26.71/0.8045 
                & 31.23/0.9176 \\


                ~ & SMFANet+~\cite{smfanet} & 496K & 28G 
                & 32.51/0.8985 
                & 28.87/\textcolor{black}{0.7872} 
                & 27.74/0.7412 
                & 26.56/0.7976 
                & 31.29/0.9163 \\
                

                ~ & MambaIR-light~\cite{mambair} & 924K & 85G 
                & 32.42/0.8977 
                & 28.74/0.7847 
                & 27.68/0.7400 
                & 26.52/0.7983 
                & 30.94/0.9135\\

                ~ & MambaIRv2-light~\cite{mambairv2} & 790K & 76G
                &  32.51/0.8992 
                & 28.84/0.7878 
                & \textcolor{blue}{27.75}/0.7426 
                & 26.82/0.8079 
                & 31.24/0.9182\\

                ~ & CATANet~\cite{catanet} & 535K & 34G
                & 32.58/0.8998
                & 28.90/0.7880
                & \textcolor{blue}{27.75}/0.7427 
                & 26.87/0.8081
                & 31.31/0.9183 \\
                
                \cline{2-9}
                ~ & \textbf{CUST-Base(Ours)} & 740K & 91G
                & \textcolor{red}{32.70/0.9008}
                & \textcolor{red}{29.00/0.7899}
                & \textcolor{red}{27.81}/\textcolor{blue}{0.7441}
                & \textcolor{red}{26.92}/\textcolor{blue}{0.8085}
                & \textcolor{blue}{31.45/0.9190} \\

                ~ & \textbf{CUST-Base+(Ours)} & 740K & 98G
                & \textcolor{blue}{32.59/0.9000}
                & \textcolor{blue}{28.94/0.7893}
                & \textcolor{red}{27.81/0.7444} 
                & \textcolor{blue}{26.91}/\textcolor{red}{0.8086} 
                & \textcolor{red}{31.48/0.9191} \\
                
				\hline
            
		\end{tabular}}
	\end{center}
\vspace{-3mm}
\end{table*}
\begin{table*}[t]
	\caption{
    Quantitative comparison of CUST-Small and CUST-Small+ with state-of-the-art efficient SISR models. The best and second-best performances are highlighted in \textcolor{red}{red} and \textcolor{blue}{blue}, respectively.}
 \vspace{-3mm}
	\label{tab:small}
	\renewcommand\arraystretch{1.1}
	\begin{center}
		\resizebox{\textwidth}{!}{
			\begin{tabular}{| c | c | c | c | c | c | c | c | c |}
				\hline
				Scale & Method & \#Params & \#FLOPs & Set5 & Set14 & B100 & Urban100 & Manga109 \\
				\hline

            
				\multirow{10}*{$\times 2$} 
                & IMDN ~\cite{IMDN}  &694K   &156G  
                & 38.00/0.9605 
                & 33.63/0.9177 
                & 32.19/0.8996 
                & 32.17/0.9283 
                & 38.88/0.9774 \\
                
				~ & LatticeNet~\cite{Lattice}    &756K   &170G   
                & 38.06/0.9607 
                & 33.70/0.9187 
                & 32.20/0.8999 
                & 32.25/0.9288  
                & 38.94/0.9774 \\
                
				~ & RFDN-L ~\cite{rfdn}  &626K   &146G  
                & 38.08/0.9606  
                & 33.67  /0.9190  
                & 32.18/0.8996 
                & 32.24/0.9290 
                & 38.95/0.9773 \\
                
				~ & SRPN-Lite~\cite{srpn}       &609K  &140G 
                &38.10/0.9608 
                & 33.70/0.9189 
                & 32.25/0.9005
                & 32.26/0.9294 
                & -  \\
                
				~ & HNCT ~\cite{hnct}       &357K     &82G 
                & 38.08/0.9608  
                & 33.65/0.9182  
                & 32.22/0.9001  
                & 32.22/0.9294  
                & 38.87/0.9774 \\
                
				~ & FMEN~\cite{fmen}  &748K     &172G  
                & 38.10/0.9609  
                & 33.75/0.9192  
                & 32.26/0.9007  
                & 32.41/0.9311  
                & 38.95/0.9778 \\

                 ~ & NGswin~\cite{ngswin}    &990K     &140G 
                &38.05/\textcolor{blue}{0.9610}
                & 33.79/0.9199 
                & 32.27/0.9008 
                &32.53/0.9324
                &38.97/0.9777 \\

                ~ & LMLT-Base~\cite{lmlt} & 652K & 158G  
                & 38.10/\textcolor{blue}{0.9610} 
                & 33.76/0.9201
                & 32.28/\textcolor{blue}{0.9012}
                & 32.52/0.9316 
                & \textcolor{red}{39.24/0.9783} \\

                \cline{2-9}
                ~ & \textbf{CUST-Small(Ours)} & 277K & 128G
                & \textcolor{red}{38.19/0.9612}
                & \textcolor{red}{33.88/0.9210}
                & \textcolor{red}{32.31/0.9014}
                & \textcolor{red}{32.73/0.9334} 
                & 39.15/0.9780 \\

                ~ & \textbf{CUST-Small+(Ours)} & 277K & 140G 
                & \textcolor{blue}{38.18}/\textcolor{red}{0.9612} 
                & \textcolor{blue}{33.80/0.9207}
                & \textcolor{blue}{32.30}/\textcolor{red}{0.9014} 
                & \textcolor{blue}{32.72/0.9333} 
                & \textcolor{blue}{39.18/0.9782} \\
                \cline{2-9}

				\hline
				\hline


				\multirow{10}*{$\times 3$} 
                & IMDN ~\cite{IMDN}         &703K     &72G 
                & 34.36/0.9270 
                & 30.32/0.8417 
                & 29.09/0.8046 
                & 28.17/0.8519
                & 33.61/0.9445 \\
                
				~ & LatticeNet~\cite{Lattice}           &765K   &76G 
                &34.40/0.9272 
                & 30.32/0.8416 
                & 29.10/0.8049
                & 28.19/0.8513 
                & 33.63/0.9442 \\
    
				~ & RFDN-L~\cite{rfdn}  &633K   &66G   
                &34.47/0.9280 
                &30.35/0.8421 
                &29.11/0.8053  
                &28.32/0.8547  
                &33.78/0.9458 \\
    
				~ & SRPN-Lite~\cite{srpn}       &615K    &63G  
                & 34.47/0.9280 
                & 30.38/0.8425 
                & 29.16/0.8061 
                & 28.22/0.8534 
                & - \\
    
				~ & HNCT ~\cite{hnct}       &363K     &38G 
                & 34.47/0.9275 
                & 30.44/0.8439 
                & 29.15/0.8067  
                & 28.28/0.8557 
                & 33.81/0.9459 \\
    
				~ & FMEN~\cite{fmen}  &757K     &77G  
                & 34.45/0.9275 
                & 30.40/0.8435  
                & 29.17/0.8063 
                & 28.33/0.8562 
                & 33.86/0.9462 \\

               
               ~ & NGswin~\cite{ngswin}           &1,007K     &67G 
               & 34.52/0.9282 
               & 30.53/0.8456 
               & 29.19/0.8078 
               & 28.52/0.8603
               & 33.89/0.9470 \\

               ~ & LMLT-Base~\cite{lmlt} & 660K & 75G  
               & 34.58/0.9285
               & \textcolor{blue}{30.53/0.8458}
               & \textcolor{blue}{29.21/0.8084}
               & 28.48/0.8581 
               & \textcolor{red}{34.18}/\textcolor{blue}{0.9477}\\

                \cline{2-9}
               ~ & \textbf{CUST-Small(Ours)} & 317K & 62G
                & \textcolor{red}{34.65/0.9291} 
                & \textcolor{blue}{30.58}/\textcolor{red}{0.8466}
                & \textcolor{red}{29.24/0.8090}
                & \textcolor{red}{28.70/0.8624}
                & \textcolor{blue}{34.17}/\textcolor{blue}{0.9477} \\

                ~ & \textbf{CUST-Small+(Ours)} & 317K & 67G
                & \textcolor{blue}{34.60/0.9289}   
                &\textcolor{red}{30.59}/\textcolor{blue}{0.8464}
                & \textcolor{blue}{29.23/0.8089}
                & \textcolor{red}{28.74/0.8630}
                & \textcolor{blue}{34.17}/\textcolor{red}{0.9479} \\
                
                \cline{2-9}

    		\hline
				\hline


    
				\multirow{10}*{$\times 4$} 
                & IMDN ~\cite{IMDN}         &715K     &41G 
                & 32.21/0.8948 
                & 28.58/0.7811 
                & 27.56/0.7353
                & 26.04/0.7838 
                & 30.45/0.9075 \\
                
				~ & LatticeNet~\cite{Lattice}           &777K   &44G   
                & 32.18/0.8943 
                & 28.61/0.7812 
                & 27.57/0.7355 
                & 26.14/0.7844
                & 30.54/0.9075 \\
    
				~ & RFDN-L ~\cite{rfdn}  &643K   &38G  
                & 32.28/0.8957 
                & 28.61/0.7818 
                & 27.58/0.7363 
                & 26.20/0.7883 
                & 30.61/0.9096 \\
    
                ~ & SRPN-Lite~\cite{srpn}       &623K    &36G 
                & 32.24/0.8958 
                & 28.69/0.7836 
                & 27.63/0.7373
                & 26.16/0.7875
                & - \\
                
				~ & HNCT ~\cite{hnct}       &373K     &22G 
                & 32.31/0.8957  
                & 28.71/0.7834 
                & 27.63/0.7381
                & 26.20/0.7896 
                & 30.70/0.9112 \\
    
				~ & FMEN~\cite{fmen}  &769K     &44G 
                & 32.24/0.8955  
                & 28.70/0.7839
                & 27.63/0.7379 
                & 26.28/0.7908 
                & 30.70/0.9107 \\

                ~ & NGswin~\cite{ngswin}           &1,019K     &36G 
                & 32.33/0.8963 
                & 28.78/0.7859
                & 27.66/0.7396
                & 26.45/0.7963
                & 30.80/0.9128 \\

                ~ & LMLT-Base~\cite{lmlt} & 672K & 41G 
                & 32.38/0.8971
                & \textcolor{blue}{28.79}/0.7859
                & \textcolor{blue}{27.70}/0.7403
                & \textcolor{blue}{26.44}/0.7947 
                & 31.09/0.9139 \\

                \cline{2-9}
                
                ~ & \textbf{CUST-Small(Ours)} & 309K & 42G
                & \textcolor{red}{32.46/0.8982} 
                & \textcolor{red}{28.85}/\textcolor{blue}{0.7862}
                & \textcolor{red}{27.73/0.7411}
                & \textcolor{red}{26.60/0.7995}
                & \textcolor{red}{31.14/0.9145} \\

                ~ & \textbf{CUST-Small+(Ours)} & 309K & 45G
                & \textcolor{blue}{32.44/0.8980}
                & \textcolor{red}{28.85/0.7864}
                & \textcolor{red}{27.73}/\textcolor{blue}{0.7409}
                & \textcolor{red}{26.60}/\textcolor{blue}{0.7989}
                & \textcolor{blue}{31.12/0.9143}\\
                
                \cline{2-9}
				\hline
            
		\end{tabular}}
	\end{center}
\vspace{-3mm}
\end{table*}
\begin{table}[ht]
    \centering
    \caption{
     Efficiency comparison between CUST-Base+ and state-of-the-art models. All evaluations are conducted on an NVIDIA RTX 3090 GPU. SSM-based models are marked with $\dagger$, while the others are ViT-based models.}
    \label{tab:memtime}
    \resizebox{0.8\linewidth}{!}{
        \begin{tabular}{l | cc | cc | cc}
            \toprule
            \multirow{2}{*}{Model} & \multicolumn{2}{c|}{Scale $\times 2$} & \multicolumn{2}{c|}{Scale $\times 3$} & \multicolumn{2}{c}{Scale $\times 4$} \\
            \cmidrule{2-7}
            & Mem(M) & Time(ms) & Mem(M) & Time(ms) & Mem(M) & Time(ms) \\
            \midrule
            SwinIR-light~\cite{swinir} & 1286.76 & 920.79 & 595.74 & 308.76 & 350.57 & 195.89 \\
            SwinIR-NG~\cite{ngswin} & 1236.14 & 984.24 & 572.85 & 365.89 & 337.68 & 237.01 \\
            SRFormer-light~\cite{srformer} & 1184.28 & 976.84 & 537.54 & 330.32 & 329.08 & 197.18 \\
            SPIN~\cite{spin} & 5799.40 & 3190.69 & 1200.02 & 1296.61 & 456.64 & 702.44 \\
            HIT-SIR~\cite{hit} & 1803.97 & 441.74 & 1464.10 & 217.60 & 1331.16 & 143.74 \\
            MambaIR-light~\cite{mambair} $\dagger$ & 1695.43 & 600.75 & 761.06 & 235.42 & 438.25 & 130.08 \\
            MambaIRv2-light~\cite{mambairv2} $\dagger$ & 2823.96 & 922.30 & 1250.76 & 414.67 & 748.42 & 241.83 \\
            CATANet~\cite{catanet} & 6935.17 & 678.28 &3374.99 & 238.62 & 1818.34 & 144.89 \\
            \midrule
            


            \textbf{CUST-Base$+$(Ours)} 
            & \textbf{1211.89} & \textbf{613.02}
            & \textbf{540.49} & \textbf{243.53}
            & \textbf{328.17} & \textbf{146.04} \\
            
            \bottomrule
        \end{tabular}
    }
\end{table}

\paragraph{Datasets.} Following previous works~\cite{hnct, shufflemixer, Lattice}, we utilize the DIV2K~\cite{div2k} dataset for training. The DIV2K dataset consists of 800 training images and 100 validation images, and is widely adopted as a standard benchmark for SISR tasks. For performance evaluation, we employ five standard benchmark datasets: Set5~\cite{set5}, Set14~\cite{Set14}, BSD100~\cite{bsd100}, Urban100~\cite{urban100}, and Manga109~\cite{manga109}.

\paragraph{Model Implementation Details.} We design four variants of our model: CUST-Base, CUST-Base+, CUST-Small and CUST-Small+. CUST-Base is configured with 40 channels and 12 blocks, while CUST-Small consists of 30 channels and 8 blocks. For both models, the window sizes in the MEDA modules are cyclically assigned from the set $[12, 14, 16, 18]$ across the consecutive blocks. CUST-Base+ and CUST-Small+ shares the same channel and block configuration as CUST-Base and CUST-Small but utilizes a uniform window size of 18 for all blocks. All models are trained for $5 \times 10^5$ iterations with a batch size of 32. We utilize the Adam optimizer~\cite{Adam} with $\beta_1=0.9$ and $\beta_2=0.99$. For data augmentation, random horizontal flips and $90^\circ$ rotations are applied. The input patch size for training is fixed at $64 \times 64$. The initial learning rate is set to $5 \times 10^{-4}$ with a warm-up period of 20,000 iterations. Subsequently, the learning rate is managed using the MultiStepLR scheduler, which halves the learning rate at iterations 250,000, 400,000, 450,000, and 475,000. Gradient clipping is applied with a threshold of 0.1.

\paragraph{Evaluation Metrics.} To evaluate the quality of the super-resolved images, we employ Peak Signal-to-Noise Ratio (PSNR) and Structural Similarity Index (SSIM)~\cite{SSIM} as the primary metrics. Furthermore, to assess the efficiency of the SR models, we measure GPU memory consumption and inference latency. For inference time, we report the average runtime calculated over 50 randomly selected images with resolutions of $640 \times 360$, $427 \times 240$, and $320 \times 180$ corresponding to $\times 2$, $\times 3$, and $\times 4$ scaling factors, respectively. Peak GPU memory consumption is monitored using \texttt{torch.cuda.max\_memory\_allocated()} in PyTorch.

\subsection{Comparisons with State-of-the-Art Methods}
\paragraph{Image Reconstruction Comparisons.}



The comparisons of our proposed CUST with other lightweight or efficient SR models are presented in Table~\ref{tab:base} and Table~\ref{tab:small}. We compare the CUST-Base and CUST-Base+ models against a broad range of competitors, including SwinIR-light~\cite{swinir}, ELAN-light~\cite{elan}, HPI-Net~\cite{hpinet}, SwinIR-NG~\cite{ngswin}, SRFormer-light~\cite{srformer}, SPIN~\cite{spin}, OSFFNet~\cite{osffnet}, HIT-SIR~\cite{hit}, SMFANet+~\cite{smfanet}, MambaIR-light~\cite{mambair}, MambaIRv2-light~\cite{mambairv2}, and CATANet~\cite{catanet}. To ensure a fair evaluation under lightweight constraints, all comprehensive benchmark models are constrained to those trained on the DIV2K~\cite{div2k} dataset, while variants utilizing larger joint datasets (\eg, DF2K) or heavy-weight configurations are referenced through their respective lightweight versions (\eg, SwinIR-light~\cite{swinir}). These comparison models are categorized into CNN-based~\cite{osffnet, smfanet}, Transformer-based~\cite{swinir, elan, hpinet, ngswin, srformer, spin, hit}, and State-Space Model (SSM)-based~\cite{mambair, mambairv2} architectures. For CUST-Small and CUST-Small+, we evaluate their performance against IMDN~\cite{IMDN}, LatticeNet~\cite{Lattice}, RFDN-L~\cite{rfdn}, SRPN-Lite~\cite{srpn}, HNCT~\cite{hnct}, FMEN~\cite{fmen}, NGSwin~\cite{ngswin}, and LMLT-Base~\cite{lmlt}.

The proposed CUST-Base and CUST-Base+ models demonstrate superior performance compared to existing lightweight SR models. In particular, at $\times$3 and $\times$4 scales, they achieve the highest performance on all benchmark datasets. Notably, compared to CATANet~\cite{catanet}, the second-best performing model, CUST-Base yields an average PSNR improvement of 0.068 dB and 0.094 dB at $\times$3 and $\times$4 scales, respectively. Furthermore, it consistently matches or exceeds the performance of state-of-the-art models across various datasets at $\times$2 scale, demonstrating sustained excellence.

Similarly, the proposed CUST-Small and CUST-Small+ models exhibit outstanding performance compared to other efficient SR models. Especially at $\times$4 scale, CUST-Small achieves the most superior performance on all datasets, outperforming the next-best model, LMLT-Base~\cite{lmlt}, by an average of 0.063 dB in PSNR. Beyond this, CUST-Small demonstrates consistent superiority by rivaling or surpassing existing top-performing models at $\times$2 and $\times$3 scales. These results confirm that the proposed CUST architecture is not restricted to specific magnification factors but operates robustly across various resolution conditions.

\noindent\paragraph{Memory and Running Time Comparisons.}To evaluate the efficiency of the proposed CUST, we measure the peak GPU memory footprint and inference latency, comparing it with other ViT-based models~\cite{swinir, srformer, catanet, spin, ngswin, hit} and SSM-based models~\cite{mambair, mambairv2}. As presented in Table~\ref{tab:memtime}, our model demonstrates a practical trade-off between memory usage and inference speed. Specifically, CUST-Base+ utilizes $82.8\%$ less memory on average across all scales compared to CATANet~\cite{catanet}, while maintaining competitive inference latency ($146.04$ ms vs. $144.89$ ms at scale $\times4$). Furthermore, although our model consumes $0.87\%$ more memory on average than SRFormer-light~\cite{srformer}—the most memory-efficient baseline—it achieves a $29.8\%$ faster inference speed. Compared to other ViT-based models~\cite{swinir, ngswin, spin} and the SSM-based MambaIRv2~\cite{mambairv2}, CUST-Base+ consistently provides a more lightweight and faster solution. These results highlight that our proposed model strikes a favorable balance between performance and practical efficiency. To validate the scalability of our method at larger resolutions, we further analyze the performance on the Test2k~\cite{div8k} dataset in Appendix B, and provide a scaling analysis between FLOPs and inference time in Appendix C.

\noindent\paragraph{Qualitative Comparisons.}
\begin{figure*}[t]
    \centering
    \includegraphics[width=\textwidth]{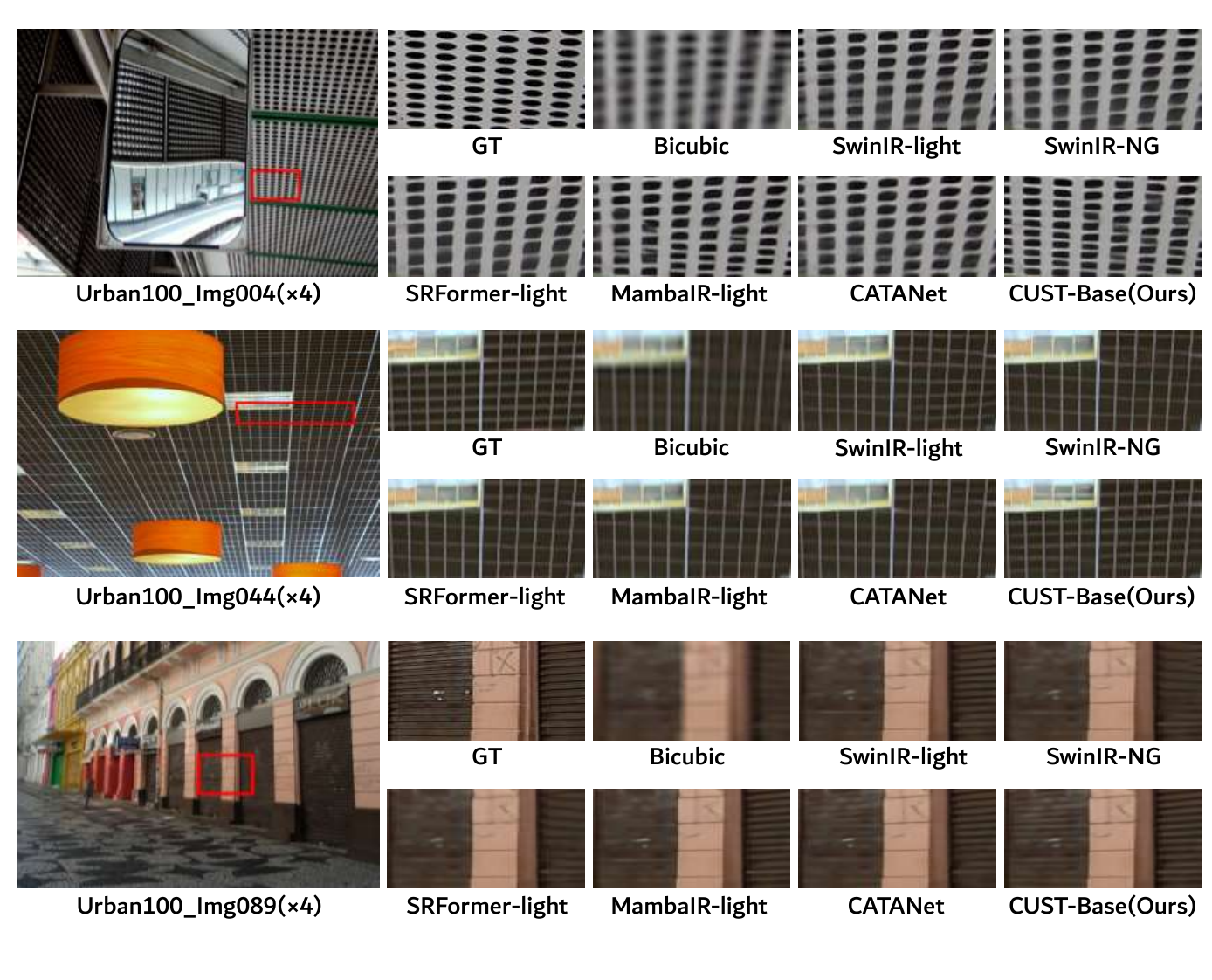} 
    \caption{
    Qualitative comparison of state-of-the-art SR models at Urban100 $\times4$ scale. Red boxes indicate the regions selected for detailed comparison.
    }
    \label{fig:qualitative}
\end{figure*}
\begin{figure*}[t]
    \centering
    \includegraphics[width=\textwidth]{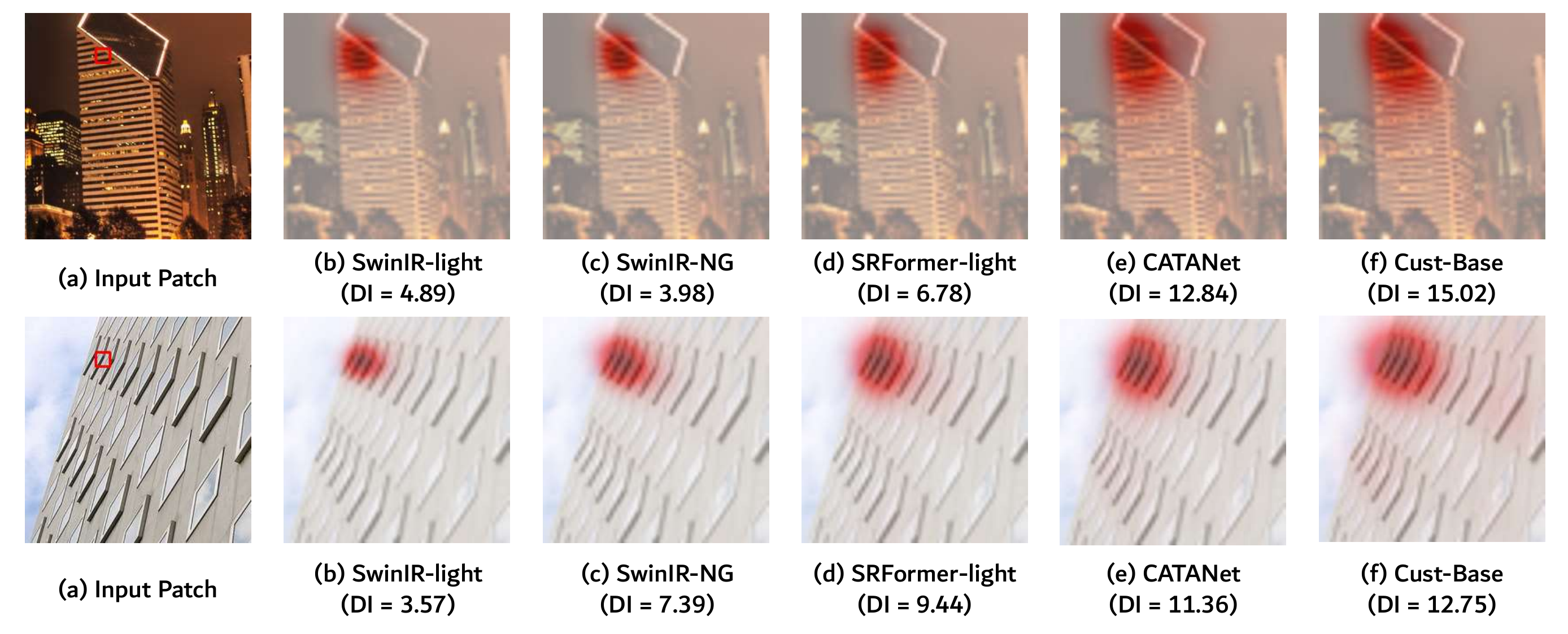} 

    \caption{
    LAM (Local Attribution Maps) comparison between the proposed model and other ViT-based lightweight SR models. As illustrated, the proposed model utilizes a broader spatial range for image reconstruction compared to other models.
    }
    \label{fig:lam_model}
\end{figure*}

Figure~\ref{fig:qualitative} illustrates the superior qualitative performance of the proposed model on the Urban100~\cite{urban100} dataset compared to other state-of-the-art models~\cite{swinir, ngswin, srformer, mambair, catanet}. As shown in the visual results, our model successfully reconstructs correct geometric orientation and structural integrity, even in dense and repetitive patterns, by accurately capturing the underlying structural context. We further analyze the receptive field of our model using Local Attribution Maps (LAM)~\cite{lam}. As illustrated in Figure~\ref{fig:lam_model}, the proposed model references a significantly broader spatial range during inference compared to other ViT-based models~\cite{swinir, srformer, ngswin, catanet}. This confirms that our architecture effectively transcends fixed window constraints to capture a wider context.


\subsection{Ablation Stuides}

\paragraph{Effects of CANA and MEDA.}
\begin{table}[t]
    \centering
    \caption{Ablation study of CANA and MEDA modules in CUST-Small for scale $\times 4$. The second and third rows show the performance drop when removing each module. The fourth and fifth rows represent networks built exclusively with a single module, scaled to match the parameter count of CUST-Small.}
    \label{tab:module_full}
    \resizebox{0.85\textwidth}{!}{
        \begin{tabular}{l|c|cc|c|c} 
            \toprule
            Module & Scaling & \#Params & \#FLOPs & Urban100 & Manga109 \\
            \midrule

            \rowcolor{gray!15} \textbf{CUST-Small (baseline)} & - 
            & 309K & 42G & \textbf{26.60/0.7995} & \textbf{31.14/0.9145} \\ 
            \hline

            Only CANA & \xmark & 227K & 29G
            & 26.31/0.7914 & 30.80/0.9118 \\ 
            Only MEDA & \xmark & 214K & 27G
            & 26.21/0.7871 & 30.61/0.9087 \\ 
            \hline

            Only CANA & \cmark & 307K & 38G 
            & 26.51/0.7966 & 30.98/0.9133\\
            Only MEDA & \cmark & 288K & 35G
            & 26.38/0.7926 & 30.81/0.9113\\
            \bottomrule
        \end{tabular}
    }
\end{table}

\begin{figure*}[t]
    \centering
    \includegraphics[width=\textwidth]{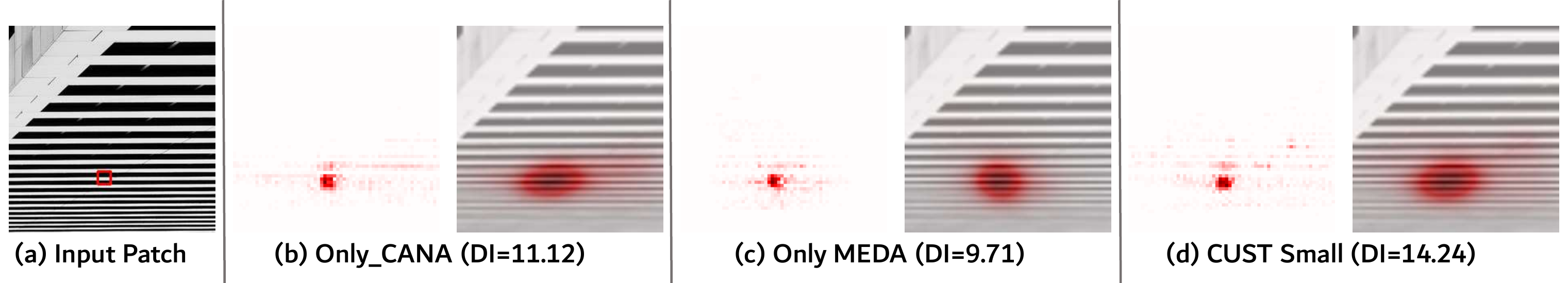} 

    \caption{
    LAM comparison between CUST-Small and its variants using only the CANA module (b) or only the MEDA module (c). The results demonstrate the synergistic effect achieved when CANA and MEDA are utilized together.
    }
    \label{fig:lam_module}
\end{figure*}

To justify the components of the proposed network, we conduct an ablation study for each module, as presented in Table~\ref{tab:module_full}. First, we compare the baseline CUST-Small (\textbf{row 1}) with variants where the MEDA and CANA modules are removed (\textbf{rows 2 and 3}). Experimental results show that removing MEDA results in a performance drop of 0.29 dB on the Urban100 dataset and 0.34 dB on the Manga109 dataset. The removal of CANA leads to even more significant degradations of 0.39 dB and 0.53 dB, respectively.

Next, to demonstrate the synergy between CANA and MEDA, we compare versions of models using only CANA or MEDA, where the number of blocks is increased to 12 to match the parameter count of CUST-Small (\textbf{rows 4 and 5}). Our experimental results show that simultaneously utilizing both modules yields a significant synergetic effect compared to using either one in isolation. Specifically, the CANA-only variant exhibits a decrease in PSNR of 0.09 dB and 0.16 dB on Urban100 and Manga109, respectively, compared to the baseline. Similarly, the MEDA-only variant shows a drop of 0.22 dB and 0.33 dB, with SSIM following a consistent trend. The LAM visualization in Figure~\ref{fig:lam_module} reveals that CUST-Small utilizes a broader receptive field than the single-module variants. This underscores the importance of capturing long-range dependencies via CANA followed by local detail refinement through MEDA.

\paragraph{Effects of Key Components in CANA and MEDA.}
\begin{table}[t]
    \centering
    \caption{Ablation study on the core components of CUST-Small on Urban100 ($\times$4).}
    \label{tab:rebut_ablation}
    \resizebox{\textwidth}{!}{ 
    \begin{tabular}{llccc}
        \toprule
        Component & Method / Variant & \#Params (K) & \#FLOPs (G) & Urban100 (PSNR/SSIM) \\
        \midrule
        \multirow{3}{*}{CANA} 
        & SwinAttn instead of CANA & 313K & 35G & 26.53/0.7975 \\
        & w/o Expansion & 309K & 39G & 26.58/0.7985 \\
        & Reduced group size ($10 \rightarrow 5$) & 309K & 42G & 26.59/0.7987 \\
        
        \midrule

        \multirow{1}{*}{MEDA} 
        & w/o Refiner \& Gate & 296K & 41G & 26.58/0.7982 \\
        \midrule
        \textbf{Baseline} & \textbf{CUST-Small (Full)} & \textbf{309K} & \textbf{42G} & \textbf{26.60/0.7995} \\
        \bottomrule
    \end{tabular}
    }
\end{table}
\begin{figure}
	\centering
	\includegraphics[width=0.8\textwidth]{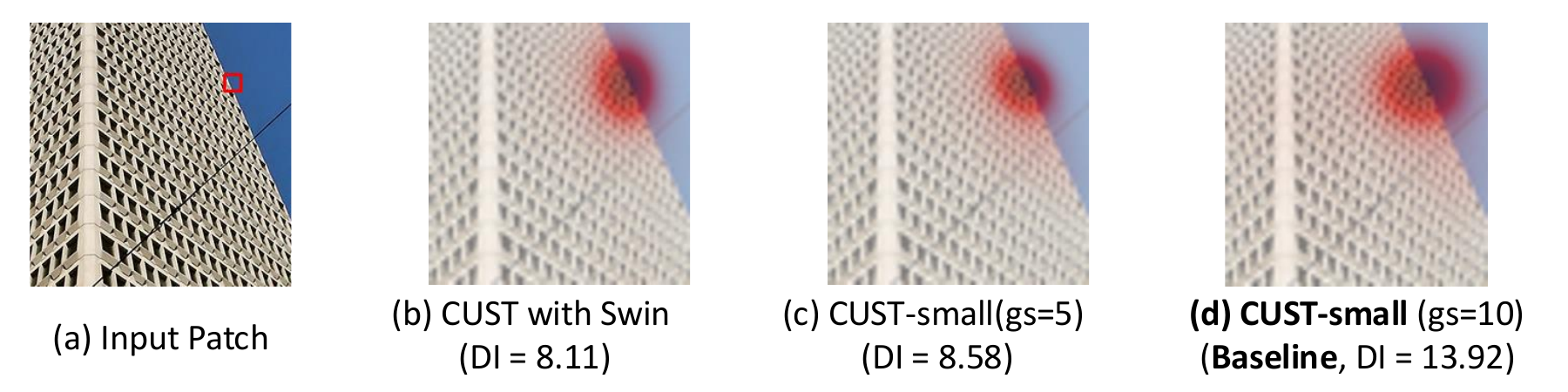}
	\caption{Visual analysis using Local Attribution Map (LAM). (a) A query patch from Urban100. (b)-(d) LAM results for different configurations. Our baseline (gs=10) shows a significantly higher Diffusion Index (DI = 13.92) compared to the Swin-based counterpart (DI = 8.11), indicating that our model effectively utilizes a broader range of spatial information for high-quality image reconstruction.}  
	\label{fig:rebut_swinlam}
\end{figure}


We verify the structural soundness and efficacy of CANA and MEDA. Table~\ref{tab:rebut_ablation} summarizes the performance variations when replacing or removing key components within each module. First, for CANA, replacing our CANA with standard Swin Attention~\cite{swin} leads to a performance drop from 26.60 dB to 26.53 dB. Moreover, scaling down search region scale factor $gs = GS/ws$, which represents the number of windows along one dimension of the search region, from the default value of $10$ (corresponding to a physical grid size of $GS=80$ under $ws=8$) to $5$ ($GS=40$), or entirely disabling the Key-Value (KV) Expansion mechanism, degrades the restoration quality, reducing the SSIM by 0.0008 and 0.0010, respectively. As illustrated by the Local Attribution Map (LAM)~\cite{lam} results in Figure~\ref{fig:rebut_swinlam}, replacing our attention mechanism or altering the group size forces the model to utilize a narrower range of input patches compared to the baseline (CUST-Small). This visual evidence proves that our semantic clustering successfully captures long-range dependencies that are otherwise overlooked by fixed local windows. Next, for MEDA, removing the error-refinement process leads to a decrease of 0.0013 (0.7995 vs. 0.7982) in SSIM, confirming that our explicit error guidance is vital for reconstructing intricate high-frequency details. Additional detailed visualizations of patch clustering in CANA and the high-frequency guidance provided by MEDA are presented in Appendix D.

\paragraph{Effects of Multi-Frequency Error-driven Calculation.}
\begin{table}

	\caption{Performance comparison of CUST-Small with varying window sizes (WS) and the Multi-Frequency Algorithm (MFA, Algorithm 1: Steps 1-5) at scale $\times$4 on Urban100.}
	\label{tab:rebut_mfa}
	\renewcommand\arraystretch{1.1}
	\begin{center}
		\resizebox{0.8\textwidth}{!}{
			\begin{tabular}{| c | c | c | c | c | }
				\hline
				Method & Window size & \#GPU Mem [M] 
                & \#FLOPs & Urban100 (PSNR/SSIM)  \\
				\hline
    
                \rowcolor{gray!15} WS +2 w/ MFA & [14,16,18,20] 
                & \textbf{314.6M} & \textbf{44G} & \textbf{26.64/0.8002} \\

                WS +4 w/o MFA &  [16,18,20,22] & 343.2M & 45G & 26.62/0.7994 \\

                \hline  
                
                \rowcolor{gray!15} \textbf{CUST-Small (baseline)} & [12,14,16,18] & \textbf{291.6M} & \textbf{42G} 
                & 26.60/\textbf{0.7995} \\

                WS +2 w/o MFA & [14,16,18,20] & 307.7M &43G & \textbf{26.61}/0.7989 \\
                \hline 

                \rowcolor{gray!15} WS -2 w/ MFA & [10,12,14,16] & \textbf{248.5M} & \textbf{40G} & \textbf{26.60/0.7989} \\
                 
                No size change w/o MFA & [12,14,16,18] & 284.9M & 
                41G & 26.58/0.7983 \\

    		\hline    

		\end{tabular}}
	\end{center}
\end{table}

\label{exp:lth}

To evaluate the efficiency of our Multi-Frequency Error-driven Algorithm (MFA, Algorithm~\ref{algo:meda}: Steps 1-5), we compare CUST-Small against variants that exclude this module across various window sizes leaving only the Overlap Self-Attention (Table~\ref{tab:rebut_mfa}). Specifically, the odd-indexed rows represent variants equipped with MFA under downscaled window configurations, whereas the even-indexed rows denote counterparts without MFA under upscaled window environments to set comparative upper bounds.

Experimental results confirm that our MFA effectively enhances reconstruction performance beyond restricted window boundaries, consequently contributing to a reduction in GPU memory overhead. As summarized in Table~\ref{tab:rebut_mfa}, integrating MFA consistently yields superior performance-efficiency trade-offs under varying window configurations. Remarkably, variants with smaller windows but equipped with MFA consistently outperform versions with a window size larger by 2, while maintaining a lower GPU memory footprint. For instance, \textit{WS +2 w/ MFA} (\textbf{row 1}) achieves an 8.3\% lower GPU memory usage alongside a higher PSNR compared to \textit{WS +4 w/o MFA} (\textbf{row 2}) (26.64 dB vs. 26.62 dB). Furthermore, a more aggressive reduction layout, such as \textit{WS -2 w/ MFA} (\textbf{row 5}), achieves a highly comparable PSNR of 26.60 dB and SSIM of 0.7989 against the \textit{WS +2 w/o MFA} (\textbf{row 4}) variant (26.61 dB / 0.7989), while drastically reducing the peak GPU memory footprint by 19.2\% (248.5M vs. 307.7M). These findings explicitly demonstrate that the multi-frequency algorithm not only compensates for the inherent spatial constraints of window-based architectures but also drastically improves practical hardware memory efficiency. A more extensive comparative analysis regarding the trade-offs between window size reduction and the integration of our proposed algorithm is provided in Appendix E.

\section{Conclusion and Limitation}

In this paper, we propose the CUST (Clustered Unit-level Similarity Transformer) for lightweight image super-resolution. By grouping windows into search regions and clustering patches based on their affinity to window tokens, CUST effectively captures long-range dependencies while maintaining high computational efficiency. Furthermore, our MEDA module explicitly guides the reconstruction of high-frequency details through multi-scale error signals, expanding the receptive field without an excessive memory burden. Experimental results confirm that CUST achieves a superior balance between restoration quality, inference latency, and memory footprint compared to existing WSA and super-token-based models.

Despite these contributions, our work has certain limitations. First, while CANA effectively captures global context, its clustering efficacy may diminish in regions with highly stochastic or irregular textures where patch-to-window affinity becomes ambiguous. Second, although CUST is optimized for SISR, its generalizability to other low-level vision tasks—such as deblurring and denoising—remains to be fully explored. Future work will focus on enhancing the model's robustness across a wider range of degradation types and diverse vision tasks to further validate its practical utility for real-world hardware deployment.


%
%
\bibliographystyle{splncs04}
\bibliography{main}

\clearpage
\appendix

\renewcommand{\thetable}{A\arabic{table}}
\setcounter{table}{0}
\renewcommand{\thefigure}{A\arabic{figure}}
\setcounter{figure}{0}

\section{Additional Comparisons}
\label{app:qualitative}
\begin{table*}[t!]
	\caption{
    Quantitative comparison of CUST-Base and CUST-Base+ with state-of-the-art lightweight SISR models. The best and second-best performances are highlighted in \textcolor{red}{red} and \textcolor{blue}{blue}, respectively.}
	\label{tab:app_base}
	\renewcommand\arraystretch{1.1}
	\begin{center}
		\resizebox{\textwidth}{!}{
			\begin{tabular}{| c | c | c | c | c | c | c | c | c |}
				\hline
				Scale & Method & \#Params & \#FLOPs & Set5 & Set14 & B100 & Urban100 & Manga109 \\
				\hline

                \multirow{7}*{$\times 2$}

                 ~ & ESRT        & 751K & - 
                &38.03/0.9600
                &33.75/0.9184
                &32.25/0.9001
                &32.58/0.9318
                &39.12/0.9774\\

                ~ & MambaIR-light & 847K & 227G 
                &  38.13/0.9610 
                & 33.95/0.9208 
                & 32.31/0.9013 
                & 32.85/0.9349 
                & 39.20/0.9782 \\

                ~ & MambaIRv2-light & 774K & 286G
                & 38.26/\textcolor{blue}{0.9615}
                & \textcolor{blue}{34.09/0.9221} 
                & 32.36/0.9019 
                & \textcolor{red}{33.26}/\textcolor{red}{0.9378}
                & 39.35/\textcolor{red}{0.9785} \\

                ~ & CATANet & 477K & 135G
                & 38.28/\textcolor{red}{0.9617}    
                & 33.99/0.9217
                & \textcolor{blue}{32.37}/0.9023
                & 33.09/0.9372
                &  39.37/\textcolor{blue}{0.9784} \\ 

                ~ & MaIR-Small & 1355K & 542G 
                & 38.20/0.9611
                & 33.91/0.9209 
                & 32.34/0.9016
                & 32.97/0.9359 
                & 39.32/0.9779 \\

                \cline{2-9}
                ~ & \textbf{CUST-Base(Ours)} & 682K & 291G
                & \textcolor{blue}{38.31}/\textcolor{red}{0.9617}
                & 34.08/\textcolor{blue}{0.9221} 
                &  \textcolor{red}{32.40/0.9026}
                &  \textcolor{blue}{33.25}/0.9375
                & \textcolor{blue}{39.47}/\textcolor{red}{0.9785}\\

                ~ & \textbf{CUST-Base+(Ours)} & 682K & 314G
                & \textcolor{red}{38.32/0.9617} 
                & \textcolor{red}{34.10/0.9228}
                & \textcolor{red}{32.40}/\textcolor{blue}{0.9024}
                & 33.21/\textcolor{blue}{0.9377}
                &  \textcolor{red}{39.49}/\textcolor{red}{0.9785}\\
                
				\hline
				\hline


                \multirow{7}*{$\times 3$}

                ~ & ESRT    & 751K & - 
                &34.42/0.9268
                &30.43/0.8433
                &29.15/0.8063
                &28.46/0.8574
                &33.95/0.9455\\


                ~ & MambaIR-light & 913K & 149G
                & 34.63/0.9288 
                & 30.54/0.8459 
                & 29.23/0.8084 
                & 28.70/0.8631 
                & 34.12/0.9479 \\

                ~ & MambaIRv2-light & 781K & 127G
                &  34.71/0.9298 
                & 30.68/0.8483
                & 29.26/0.8098 
                & 29.01/0.8689
                & 34.41/0.9497 \\
                
                ~ & CATANet & 550K & 60G
                & 34.75/\textcolor{blue}{0.9300}   
                & 30.67/0.8481
                & 29.28/0.8101
                & 29.04/0.8689
                & 34.40/0.9499 \\

                ~ & MaIR-Small & 1363K & 241G 
                & 34.75/\textcolor{blue}{0.9300} 
                & 30.63/0.8479 
                & 29.29/0.8103 
                & 28.92/0.8676 
                & 34.46/\textcolor{blue}{0.9497} \\

                \cline{2-9}
                ~ & \textbf{CUST-Base(Ours)} & 754K & 138G
                & \textcolor{blue}{34.76}/\textcolor{red}{0.9303} 
                & \textcolor{red}{30.73/0.8488}
                & \textcolor{red}{29.34/0.8116}
                & \textcolor{red}{29.11/0.8699}
                & \textcolor{blue}{34.54/0.9497} \\

                ~ & \textbf{CUST-Base+(Ours)} & 754K & 147G
                & \textcolor{red}{34.79/0.9303}
                & \textcolor{blue}{30.71/0.8485}
                & \textcolor{blue}{29.33/0.8114}
                & \textcolor{blue}{29.10}/\textcolor{blue}{0.8698}
                & \textcolor{red}{34.57}/\textcolor{red}{0.9500}\\
                
    		      \hline
				\hline


                \multirow{7}*{$\times 4$} 

                ~ & ESRT  & 751K 
                & - 
                &32.19/0.8947
                &28.69/0.7833
                &27.69/0.7379
                &26.39/0.7962
                &30.75/0.9100\\



                ~ & MambaIR-light & 924K & 85G 
                & 32.42/0.8977 
                & 28.74/0.7847 
                & 27.68/0.7400 
                & 26.52/0.7983 
                & 30.94/0.9135\\

                ~ & MambaIRv2-light & 790K & 76G
                &  32.51/0.8992 
                & 28.84/0.7878 
                & 27.75/0.7426 
                & 26.82/0.8079 
                & 31.24/0.9182\\

                ~ & CATANet & 535K & 34G
                & 32.58/0.8998
                & 28.90/0.7880
                & 27.75/0.7427 
                & 26.87/0.8081
                & 31.31/0.9183 \\

                ~ & MaIR-Small & 1374K & 137G 
                & \textcolor{blue}{32.62}/0.8998 
                & 28.90/0.7882 
                & \textcolor{blue}{27.77}/0.7431
                & 26.73/0.8049 
                & 31.34/0.9183 \\

                \cline{2-9}
                ~ & \textbf{CUST-Base(Ours)} & 740K & 91G
                & \textcolor{red}{32.70/0.9008}
                & \textcolor{red}{29.00/0.7899}
                & \textcolor{red}{27.81}/\textcolor{blue}{0.7441}
                & \textcolor{red}{26.92}/\textcolor{blue}{0.8085}
                & \textcolor{blue}{31.45/0.9190} \\

                ~ & \textbf{CUST-Base+(Ours)} & 740K & 98G
                & 32.59/\textcolor{blue}{0.9000}
                & \textcolor{blue}{28.94/0.7893}
                & \textcolor{red}{27.81/0.7444} 
                & \textcolor{blue}{26.91}/\textcolor{red}{0.8086} 
                & \textcolor{red}{31.48/0.9191} \\
                
				\hline
            
		\end{tabular}}
	\end{center}
\end{table*}
\begin{table}[ht]
    \centering

    \caption{
     Efficiency comparison between CUST-Base+, CUST-Small+ and other state-of-the-arts Space State Model(SSM) based methods. All evaluations are conducted on an NVIDIA RTX 3090 GPU. SSM-based models are marked with $\dagger$.}
    \label{tab:app_memtime}
    \resizebox{0.83\linewidth}{!}{
        \begin{tabular}{l | cc | cc | cc}
            \toprule
            \multirow{2}{*}{Model} & \multicolumn{2}{c|}{Scale $\times 2$} & \multicolumn{2}{c|}{Scale $\times 3$} & \multicolumn{2}{c}{Scale $\times 4$} \\
            \cmidrule{2-7}
            & Mem(M) & Time(ms) & Mem(M) & Time(ms) & Mem(M) & Time(ms) \\
            \midrule
             MambaIR-light $\dagger$ & 1695.43 & 600.75 & 761.06 & 235.42 & 438.25 & 130.08 \\
            MambaIRv2-light $\dagger$ & 2823.96 & 922.30 & 1250.76 & 414.67 & 748.42 & 241.83 \\
            MaIR-Small$\dagger$ & 2136.7 & 1635.22 & 961.55 & 732.44 & 548.04 & 429.12 \\

           \hline
            \textbf{CUST-Base$+$(Ours)} 
            & \textbf{1211.89} & \textbf{613.02}
            & \textbf{540.49} & \textbf{243.53}
            & \textbf{328.17} & \textbf{146.04} \\

            \textbf{CUST-Small$+$(Ours)} 
            & \textbf{1093.05} & \textbf{403.37}
            & \textbf{487.87} & \textbf{173.53}
            & \textbf{291.44} & \textbf{108.05} \\
            
            \bottomrule
        \end{tabular}
    }
\end{table}
\begin{figure*}[t]
    \centering
    \includegraphics[width=\textwidth]{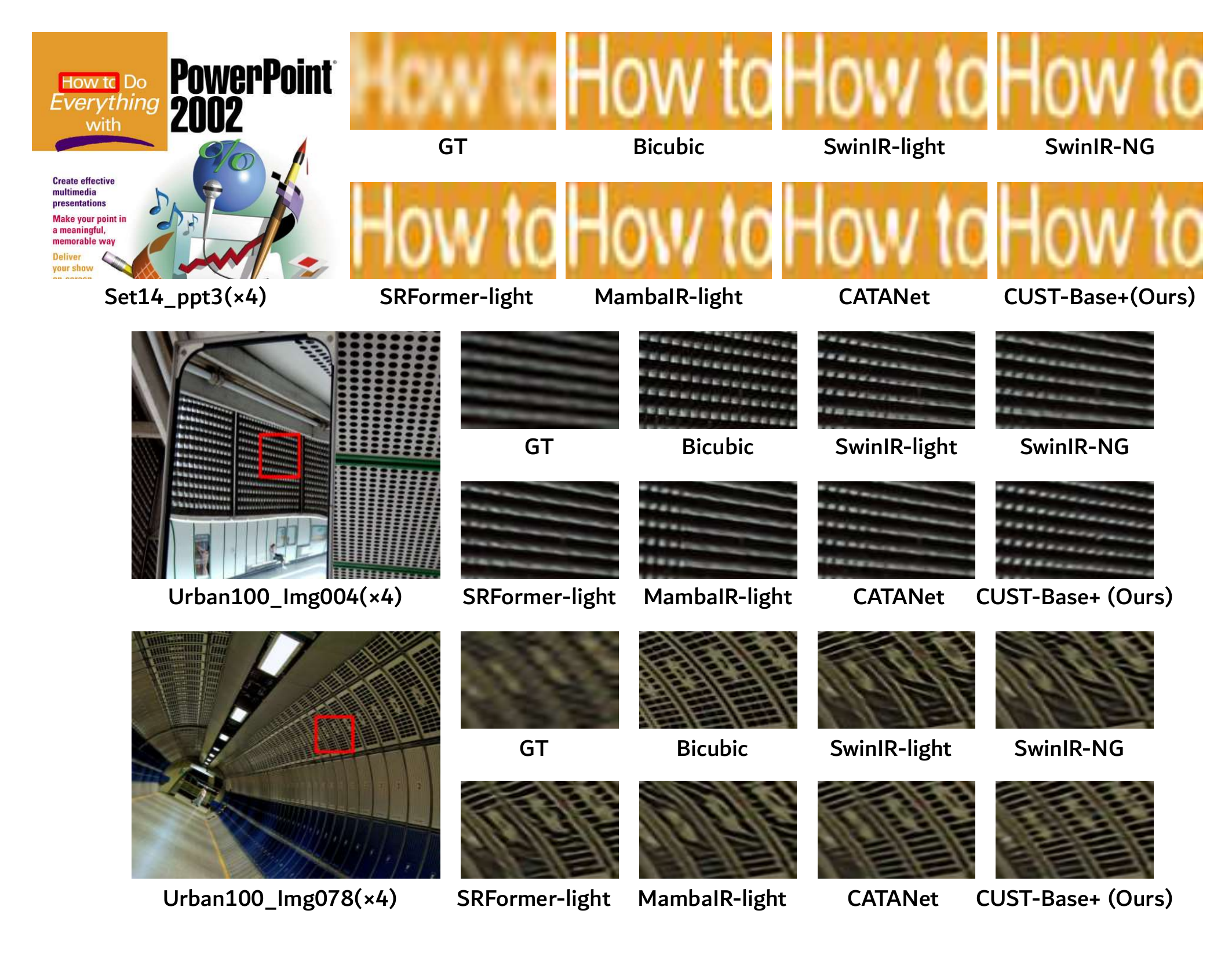} 
    \caption{
    Qualitative comparison of state-of-the-art SR models at Urban100 $\times4$ scale and Set14 $\times4$ scale. Red boxes indicate the regions selected for detailed comparison.
    }
    \label{fig:app_qualitative}
\end{figure*}

In this section, we provide additional qualitative and quantitative comparisons between CUST and state-of-the-art models. In Table~\ref{tab:app_base}, we expand the evaluations from Table~\ref{tab:base} by including MaIR-Small and ESRT alongside MambaIR-light, MambaIRv2-light, and CATANet. Furthermore, Table~\ref{tab:app_memtime} compares the peak GPU memory footprint and inference latency of CUST-Base+ against various State Space Model (SSM) based architectures, including MaIR-Small.

As shown in Table~\ref{tab:app_base}, our proposed model achieves superior performance across most scales and datasets. Notably, at the $\times3$ scale, CUST variants consistently deliver the highest reconstruction quality. Furthermore, Table~\ref{tab:app_memtime} highlights that CUST-Base+ is significantly more lightweight and faster than recent SSM-based models. Specifically, compared to MaIR-Small, our model consumes $42.40\%$ less memory and achieves a $65.09\%$ faster inference speed on average across all scales. These results demonstrate that while SSMs have emerged with a focus on efficiency, a well-designed Transformer architecture like CUST can maximize global modeling strengths while maintaining a substantially lower computational overhead.

Finally, we compare CUST-Base+ with other state-of-the-art models. As illustrated in Figure~\ref{fig:app_qualitative}, CUST-Base+ effectively captures and reconstructs structural context within repetitive patterns, mirroring the high-fidelity results of the base CUST model.

\section{Evaluation on Test2k Dataset}
\label{app:test2k}
\begin{table}
	\caption{Quantitative comparison of inference efficiency and performance on the Test2K dataset. All metrics were measured using an RTX 3090 GPU. }
	\label{tab:rebut_test2k}
	\renewcommand\arraystretch{1.1}
	\begin{center}
		\resizebox{0.8\textwidth}{!}{
			\begin{tabular}{| c | c | c | c | c | }
				\hline
				Method & \#GPU Mem [M] & \#Avg Time [ms] & Test2K (PSNR/SSIM)  \\
				\hline
    
                SwinIR-light  & 1017.63 & 372.90  & 27.72/0.7793 \\

                SwinIR-NG & 976.74 & 394.05 & 27.75/0.7803 \\
                
                SRFormer-light 
                & 944.97
                & 414.04
                & 27.75/0.7805   \\

                MambaIRv2-light~ & 2239.07 & 425.48 & 27.80/0.7820 \\

                MaIR-light & 1639.10 & 660.51 & 27.80/0.7820 \\
               
                CATANet & 6422.94 & 314.18
                & 27.81/0.7814  \\

                \hline
                \textbf{CUST-Base+(Ours)} 
                & \textbf{983.27}
                & \textbf{350.94}
                & \textbf{27.86}/\textbf{0.7828} \\
    		\hline    

		\end{tabular}}
	\end{center}
\end{table}

To evaluate the hardware efficiency of our model, we compare CUST-Base+ against state-of-the-art architectures on the Test2k dataset (average resolution of $1644 \times 1089$) in terms of restoration performance, peak GPU memory footprint, and inference latency. As shown in Table~\ref{tab:rebut_test2k}, CUST-Base+ achieves a remarkable 84.7\% GPU memory reduction compared to CATANet (983MB vs. 6.4GB) while delivering the highest PSNR of 27.86 dB. These results demonstrate that CUST effectively balances its higher theoretical FLOPs with practical inference speed by successfully mitigating memory-bandwidth bottlenecks. This justifies CUST as a robust, hardware-aware solution optimized for high-performance restoration within realistic deployment constraints.

\section{Time per Module}
\label{app:timer}
\begin{table}[ht]
\centering
\caption{Efficiency analysis with structural overhead (Others). Others is calculated as Total - (Attn + MLP), representing non-computational costs like data movement and normalization.}
\label{tab:app_moduletime}
\resizebox{0.9\textwidth}{!}{ 
\begin{tabular}{llccccccc}
\toprule
\multirow{2}{*}{Scale} & \multirow{2}{*}{Model} & \multicolumn{4}{c}{Inference Time (ms)} & \multicolumn{3}{c}{FLOPs (G)} \\
\cmidrule(lr){3-6} \cmidrule(lr){7-9}
& & Total & Attn & MLP & \textbf{Others} & Total & Attn & MLP \\
\midrule

\multirow{4}{*}{$\times 2$} 
& SwinIR-light       & 922.6 & 239.1 & 69.0  & 614.5 (67\%) & 244 & 122 & 80 \\
& SRFormer-light   & 991.8 & 203.6 & 204.6 & 583.6 (59\%) & 236 & 94  & 98 \\
& CATANet           & 678.1 & 445.7 & 124.1 & 108.3 (16\%) & 135 & 41  & 70 \\
& \textbf{CUST-Base+ (Ours)}       & \textbf{609.3} & \textbf{201.9} & \textbf{126.3} & \textbf{281.1 (46\%)} & \textbf{314} & \textbf{201} & \textbf{64} \\

\midrule
\multirow{4}{*}{$\times 3$} 
& SwinIR-light       & 319.1 & 110.6 & 31.2 & 177.3 (55\%) & 111 & 55 & 36 \\
& SRFormer-light   & 331.6 & 92.0  & 83.5 & 156.1 (47\%) & 105 & 42 & 43 \\
& CATANet         & 230.5 & 135.7 & 49.9 & 44.9 (19\%)  & 60  & 19 & 31 \\
& \textbf{CUST-Base+ (Ours)}       & \textbf{237.8} & \textbf{88.8}  & \textbf{47.9} & \textbf{101.1 (42\%)} & \textbf{147} & \textbf{89} & \textbf{21} \\

\midrule
\multirow{4}{*}{$\times 4$} 
& SwinIR-light       & 193.6 & 66.1  & 18.5 & 109.0 (56\%) & 64 & 31 & 20 \\
& SRFormer-light   & 214.1 & 58.9  & 44.2 & 111.0 (52\%) & 63 & 27 & 26 \\
& CATANet           & 138.9 & 79.1  & 25.5 & 34.3 (25\%)  & 34 & 10 & 18 \\
& \textbf{CUST-Base+ (Ours)}       & \textbf{155.6} & \textbf{59.9}  & \textbf{25.2} & \textbf{70.5 (45\%)} & \textbf{98} & \textbf{55} & \textbf{11} \\

\bottomrule
\end{tabular}
}
\end{table}

We analyze the correlation between FLOPs and inference latency in Table~\ref{tab:app_moduletime}. As observed in Table~\ref{tab:memtime}, CUST-Base+ exhibits favorable inference efficiency compared to existing ViT-based models, despite its higher FLOPs. This analysis identifies the root cause by breaking down FLOPs and latency into specific modules (Attention and MLP). The latency table comprises total inference time, attention time, MLP (FFN) time, and "Others," which represents non-computational structural overhead calculated as the total time minus the attention and MLP requirements.

Experimental results confirm that the proposed CUST significantly reduces structural overhead while performing operations optimized for GPU hardware. For instance, compared to SwinIR-light at scale $\times2$, while CUST-Base+ has approximately 79G higher attention FLOPs (201G vs. 122G), its execution time is actually 15\% faster (201.9ms vs. 239.1ms). Notably, in terms of the "Others" overhead, CUST-Base+ records 281.1ms, a 54\% reduction compared to the 614.5ms of SwinIR-light. This trend is consistently observed across SRFormer-light and other scales.

The hardware-friendly nature of CUST is further highlighted when compared to CATANet. While both models exhibit similar total inference times in Table~\ref{tab:memtime}, they differ sharply in attention efficiency. At scale $\times2$, CUST-Base+ processes approximately 160G more attention FLOPs than CATANet (201G vs. 41G) yet requires 55\% less attention latency (201.9ms vs. 445.7ms). Combined with the 83\% lower GPU memory footprint reported in Table~\ref{tab:memtime}, this proves that CUST is structured to maximize GPU processing capabilities regardless of absolute computational volume.

In conclusion, CUST overcomes structural limitations through its simplified search-region-based clustering, achieving exceptionally fast inference relative to its computational load. Note that minor discrepancies in latency between this table and Table~\ref{tab:memtime} may exist due to slight environmental fluctuations during measurement.

\section{Visualization Analyses}
\label{app:feat_vis}
\begin{figure*}[t]
    \centering
    \includegraphics[width=\textwidth]{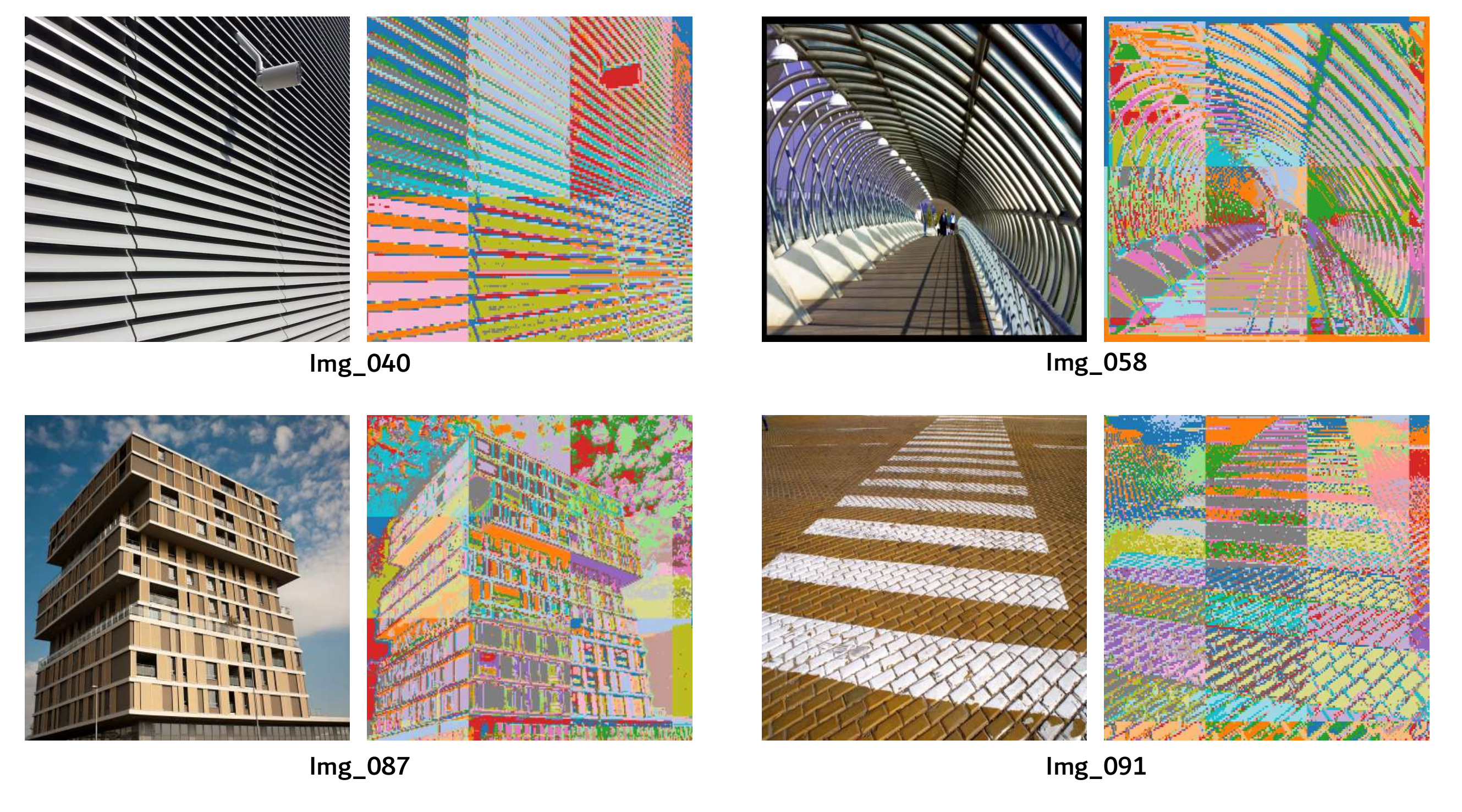} 
    \caption{
    Visualization of patch clustering in the CANA module. For each subfigure, the left image shows the original input from Urban100, and the right image visualizes clustered patches assigned the same color based on their affinity to pooled window tokens.
    }
    \label{fig:app_cana_vis}
\end{figure*}
\begin{figure*}[t]
    \centering
    \includegraphics[width=\textwidth]{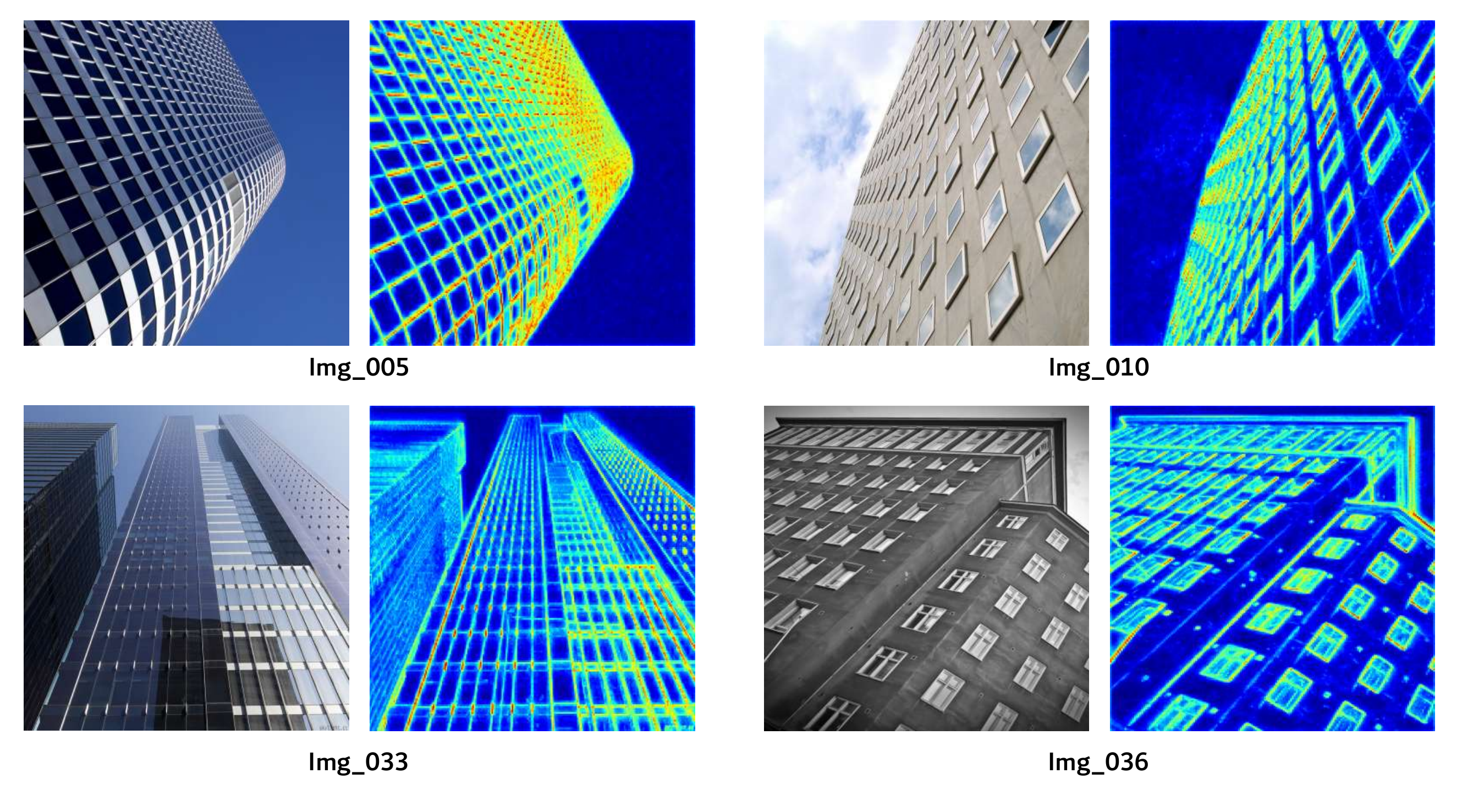} 
    \caption{
    Visualization of multi-frequency error maps in MEDA on Urban100. The right images display the high-frequency components extracted by the MFA algorithm, highlighting the module's ability to focus on structural details such as edges.
    }
    \label{fig:app_meda_vis}
\end{figure*}
\begin{figure*}[t]
    \centering
    \includegraphics[width=0.5\textwidth]{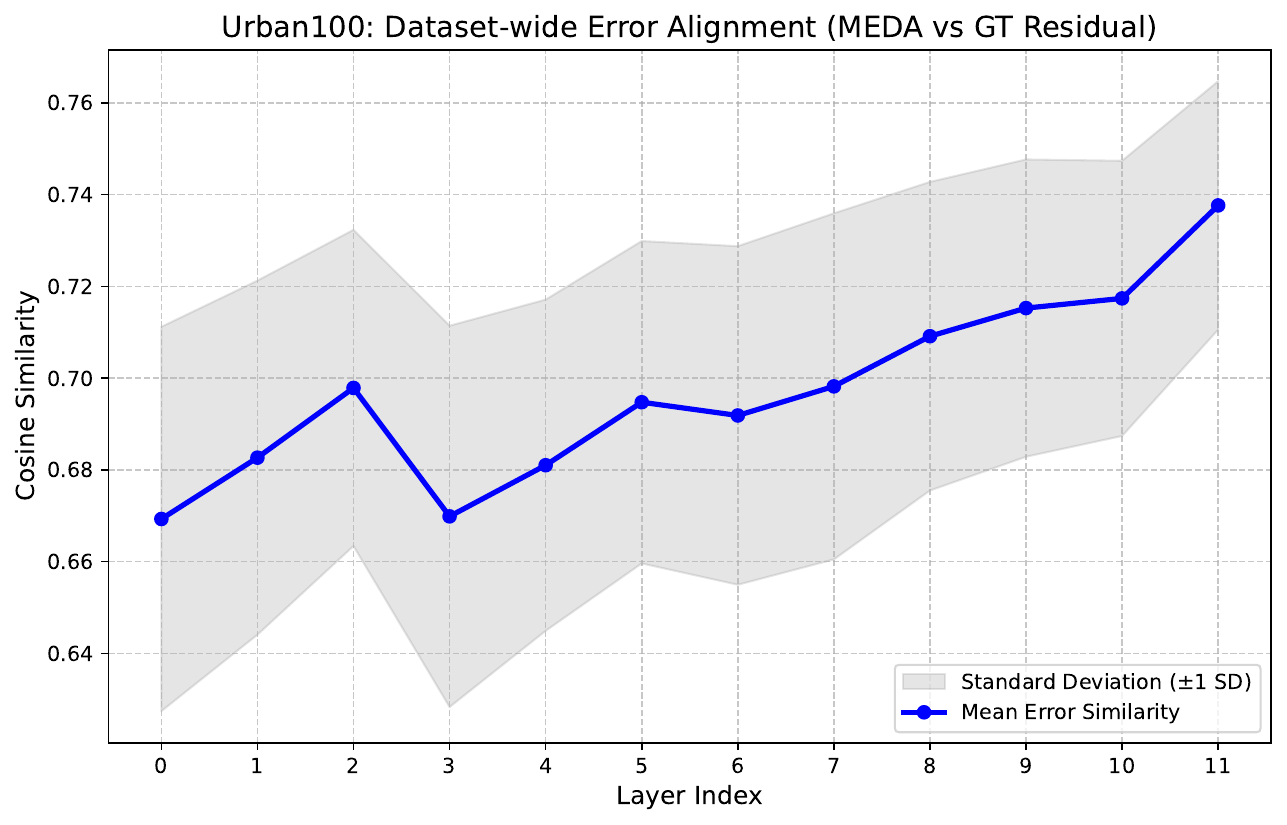} 
    \caption{
    Quantitative analysis of error alignment across layers on Urban100. The plot shows the cosine similarity between the MFA-extracted error maps and the Ground-Truth (GT) residuals ($I_{HR} - \text{Bicubic}$). The mean similarity consistently increases with depth, reaching approximately $0.74$ in the final layer, confirming that MEDA effectively identifies lost structural information.   
    }
    \label{fig:app_meda_align}
\end{figure*}

In this section, we visualize the distinctive characteristics of the CANA and MEDA modules. First, Figure~\ref{fig:app_cana_vis} illustrates how patches are grouped based on similarity within the CANA module. The right image of each subfigure visualizes patches within the search region that share the same window index, rendered in identical colors. As evident from the visualization, CANA aggregates patches with similar textures or structures even when they are spatially distant. This demonstrates its capacity to encompass significantly broader global context and model long-range dependencies compared to conventional window-based approaches.

Next, Figure~\ref{fig:app_meda_vis} visualizes the edge information extracted by the Multi-frequency Error-driven Algorithm (MFA) within MEDA. The right-side images in each section represent the primary high-frequency components identified by the algorithm, confirming that the module successfully focuses on structural information, such as edges, within the feature space.

Furthermore, we quantitatively analyze the alignment between the error maps extracted from each layer and the Ground-Truth (GT) edges in Figure~\ref{fig:app_meda_align} to verify how accurately MEDA captures the information. As defined in Equation~\ref{eq:errorgt}, the GT edge information is computed as the residual between the original HR image ($I_{HR}$) and its bicubically upsampled version:

\begin{equation}
    \label{eq:errorgt}
    GT = I_{HR} - \text{Upsize}(\text{Downsize}(I_{HR}))
\end{equation}

As shown in Figure~\ref{fig:app_meda_align}, our algorithm reflects the actual edge structures more precisely as the layers deepen. Despite the variance across the dataset (indicated by the shaded standard deviation), the mean cosine similarity shows a consistent upward trend, eventually reaching approximately $0.74$ in the final layer. This demonstrates that MEDA robustly identifies salient edge regions across diverse images, thereby inducing optimized and guided feature extraction during reconstruction.

\section{Multi-frequency Error-driven Algorithm}
\label{app:meda}
\begin{table}[t]
    \centering
    \caption{Performance comparison of CUST-Small with varying window sizes (WS) and the Multi-Frequency Algorithm (MFA) at scale $\times4$ on Urban100.}
    \label{tab:app_lth}
    \footnotesize 
    \begin{tabularx}{\textwidth}{l c c c c c}
        \toprule
        Method & Window size & \#Params & \#FLOPs & GPU Mem. & Urban100 \\
        \midrule

        WS +4 w/o MFA & [16,18,20,22] & 295K & 45G & 343.23M 
        & 26.62/0.7994 \\ 
        \rowcolor{gray!15} WS +4 w/ MFA & [16,18,20,22] & 309K & 46G & 350.13M & \textbf{26.64/0.8004} \\

        \midrule
        WS +2 w/o MFA & [14,16,18,20] & 295K & 43G & 307.72M 
        & 26.61/0.7989 \\ 
        \rowcolor{gray!15} WS +2 w/ MFA & [14,16,18,20] & 309K & 44G & 314.61M 
        & \textbf{26.64/0.8002} \\
        
        \midrule
        No size change w/o MFA & [12,14,16,18] & 295K & 41G & 284.95M & 26.58/0.7983 \\ 
        \rowcolor{gray!15} \textbf{CUST-Small (baseline)} & [12,14,16,18] & 309K & 42G & 291.64M 
        & \textbf{26.60/0.7995} \\
        
        \midrule
        WS -2 w/o MFA & [10,12,14,16] & 295K & 39G & 248.40M 
        & 26.59/0.7984 \\ 
        \rowcolor{gray!15} WS -2 w/ MFA & [10,12,14,16] & 309K & 40G & 248.48M & \textbf{26.60/0.7989} \\
        
        \midrule
        WS -4 w/o MFA & [8,10,12,14] & 295K & 38G & 248.40M 
        & 26.53/0.7973 \\ 
        \rowcolor{gray!15} WS -4 w/ MFA & [8,10,12,14] & 309K & 39G & 248.48M & \textbf{26.56/0.7983} \\
        
        \bottomrule
    \end{tabularx}
\end{table}

In this section, we extend the analysis from Section~\ref{exp:lth} to provide a deep dive into the contribution of the Multi-frequency Error-driven Algorithm (MFA) across various window size configurations for CUST-Small at $\times4$ scale. Detailed results are summarized in Table~\ref{tab:app_lth}.

Experimental results demonstrate that the proposed MFA consistently improves performance on the Urban100 dataset across all window configurations. Specifically, as the window size varies from $+4$ to $-4$, the application of MFA leads to PSNR gains of 0.02, 0.03, 0.02, 0.01, and 0.03 dB, respectively, with corresponding SSIM increases of 0.0010, 0.0013, 0.0012, 0.0005, and 0.0010. Notably, MFA exhibits superior efficiency. While the average GPU memory increase incurred by MFA is only 1.66\%, the average memory increase required for physical window expansion reaches 8.56\%. This indicates that MFA achieves higher performance gains using approximately 80.6\% fewer additional resources than the method of simply scaling up window sizes. For instance, the SSIM of \textit{WS -2 w/ MFA} (0.7989) outperforms that of the larger \textit{No size change w/o MFA} (0.7983), while simultaneously reducing the peak GPU memory footprint by 12.8\% (248.48M vs. 284.95M). Consistent with the analysis in Section~\ref{exp:lth} and Section~\ref{app:feat_vis}, these findings confirm that the proposed MFA effectively induces structural edge restoration with minimal computational overhead.

\section{Comparisons on CUST-Base and CUST-Base+}
\label{app:basevsplus}
\begin{figure}[t]
    \centering

    \begin{subfigure}{0.48\textwidth}
        \centering
        \includegraphics[width=\textwidth]{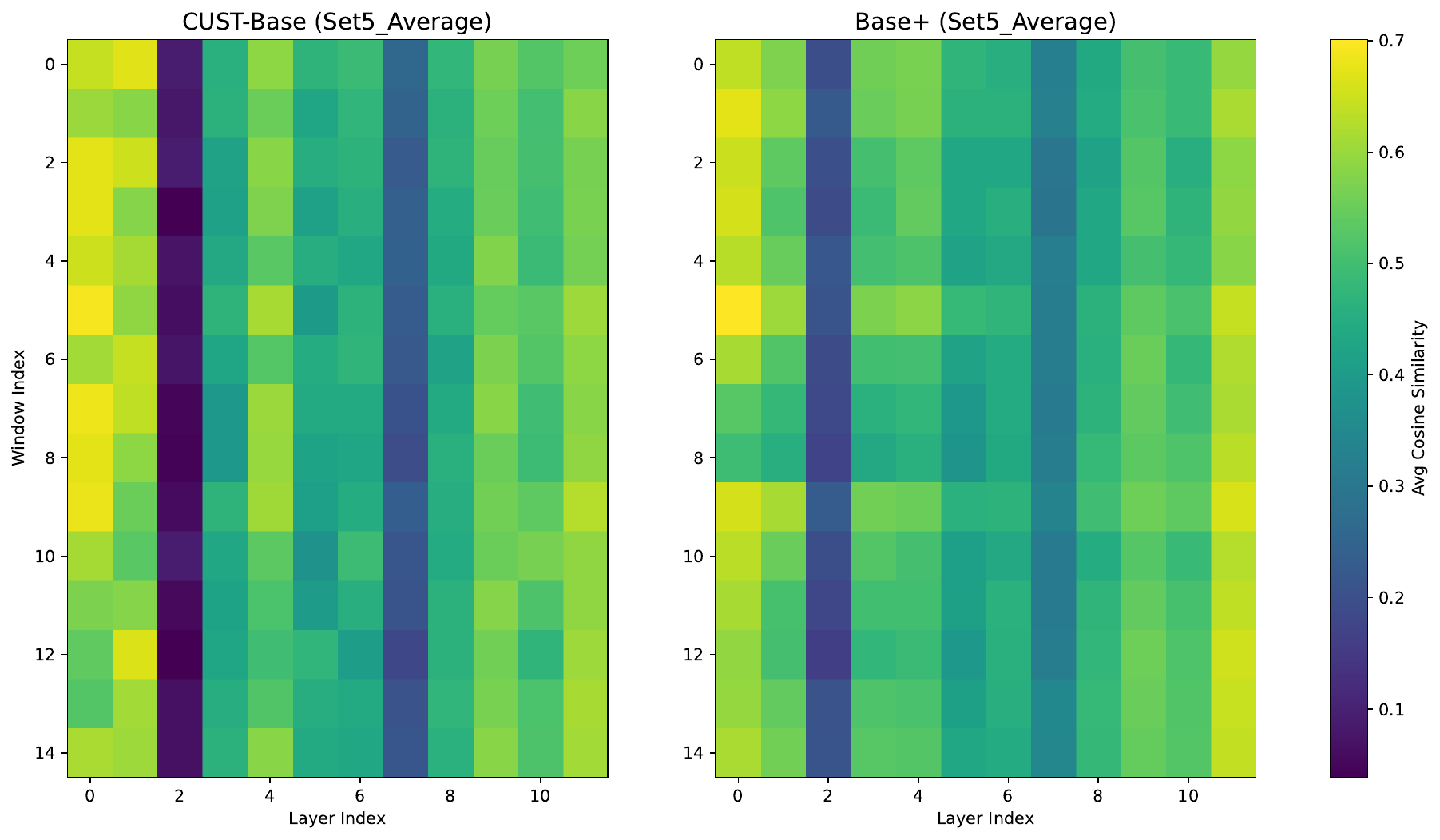}
        \caption{Intra-window patch similarity on Set5.}
        \label{fig:winsim_set5}
    \end{subfigure}
    \hfill
    \begin{subfigure}{0.48\textwidth}
        \centering
        \includegraphics[width=\textwidth]{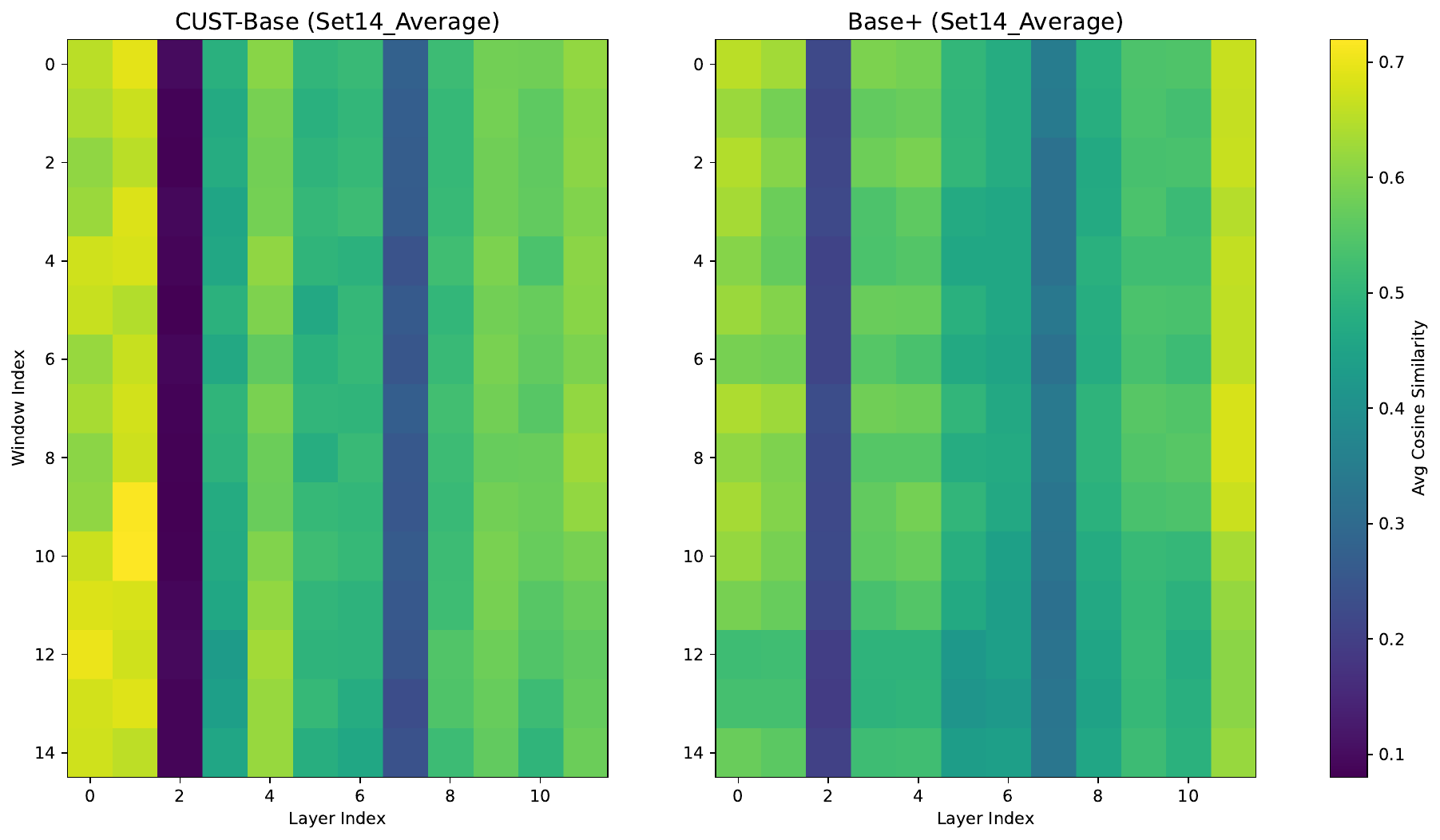}
        \caption{Intra-window patch similarity on Set14.}
        \label{fig:winsim_set14}
    \end{subfigure}

    \vspace{1em}

    \begin{subfigure}{0.48\textwidth}
        \centering
        \includegraphics[width=\textwidth]{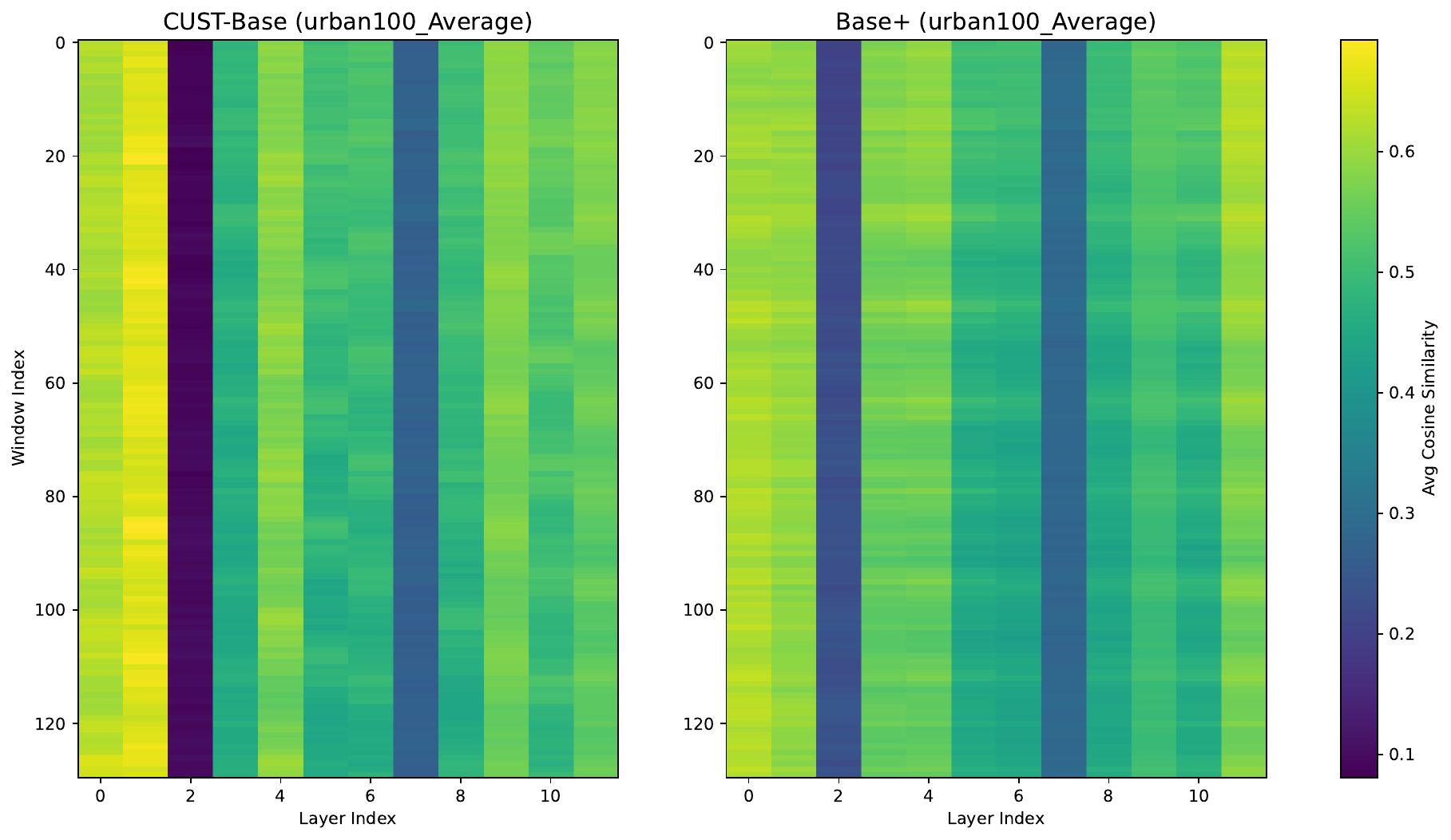}
        \caption{Intra-window patch similarity on Urban100.}
        \label{fig:winsim_urban}
    \end{subfigure}
    \hfill
    \begin{subfigure}{0.48\textwidth}
        \centering
        \includegraphics[width=\textwidth]{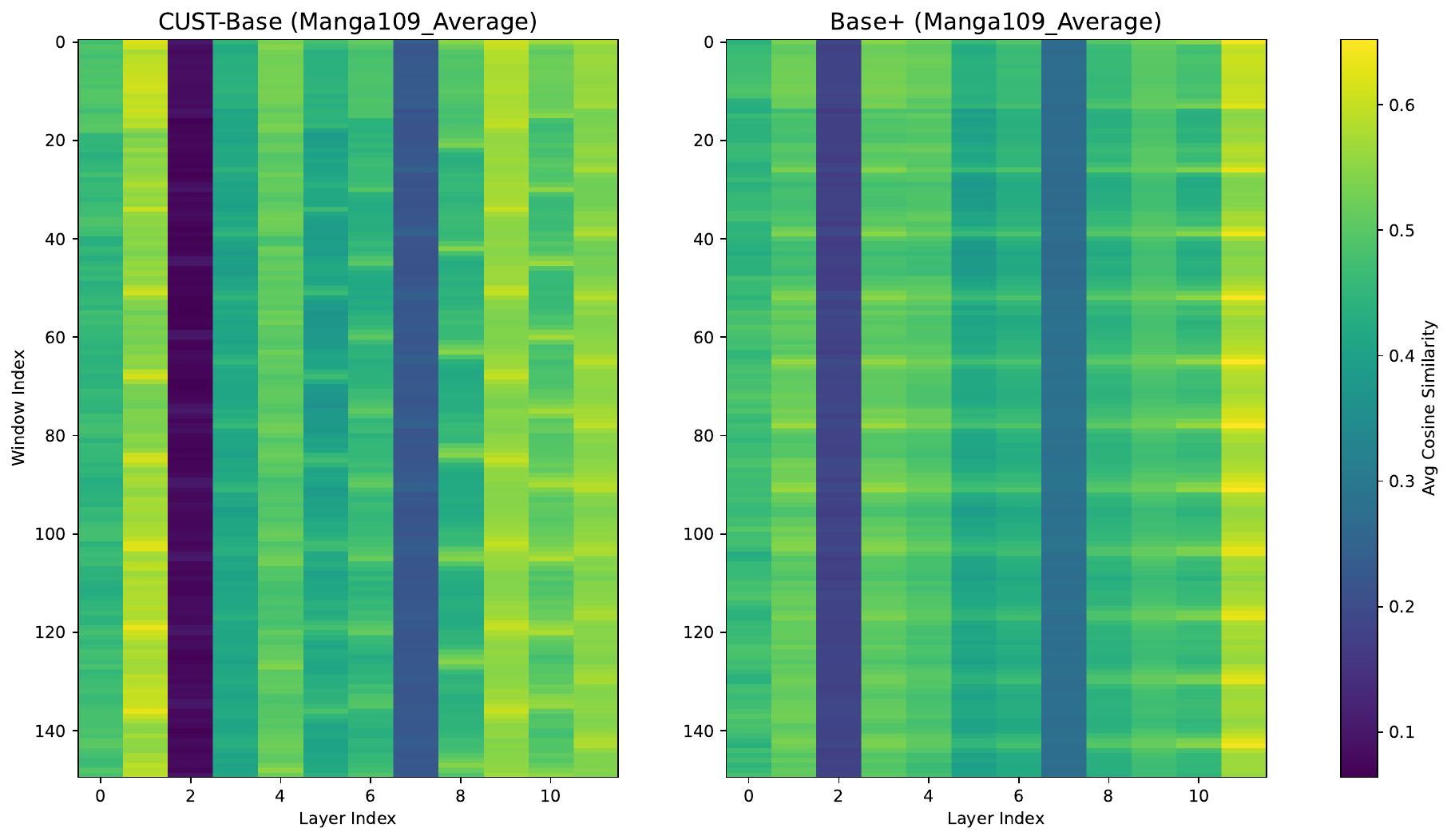}
        \caption{Intra-window patch similarity on Manga109.}
        \label{fig:winsim_manga}
    \end{subfigure}

    \caption{Analysis of intra-window patch similarity across various datasets. Each subfigure compares CUST-Base (left) and CUST-Base+ (right). The x-axis denotes the layer index, and the y-axis represents the window index, with the number of windows being 15 (Set5), 15 (Set14), 130 (Urban100), and 150 (Manga109), respectively. The color intensity indicates the mean cosine similarity between patches within each window.}
    \label{fig:app_win_sim}
\end{figure}

\begin{figure}[t]
    \centering
    \begin{subfigure}{0.48\textwidth}
        \centering
        \includegraphics[width=\textwidth]{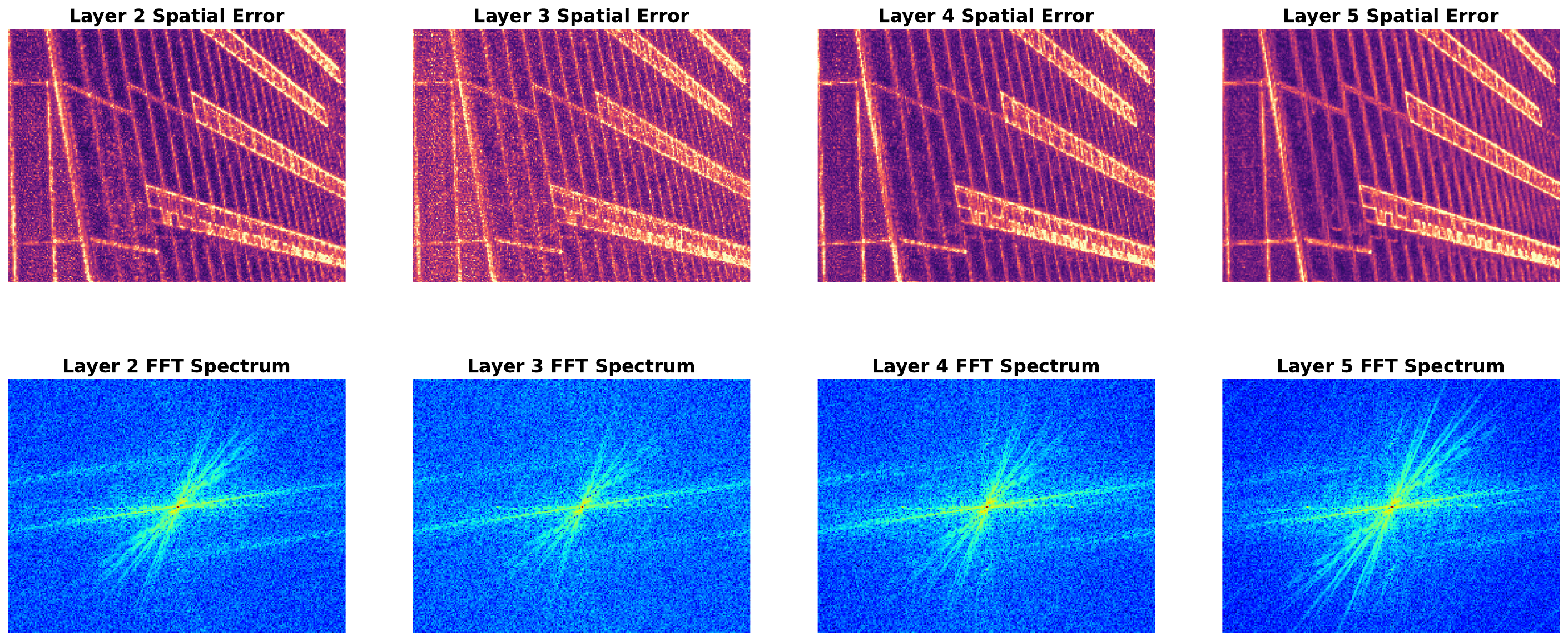}
        \caption{Urban100 img\_026}
        \label{fig:fft_1}
    \end{subfigure}
    \hfill
    \begin{subfigure}{0.48\textwidth}
        \centering
        \includegraphics[width=\textwidth]{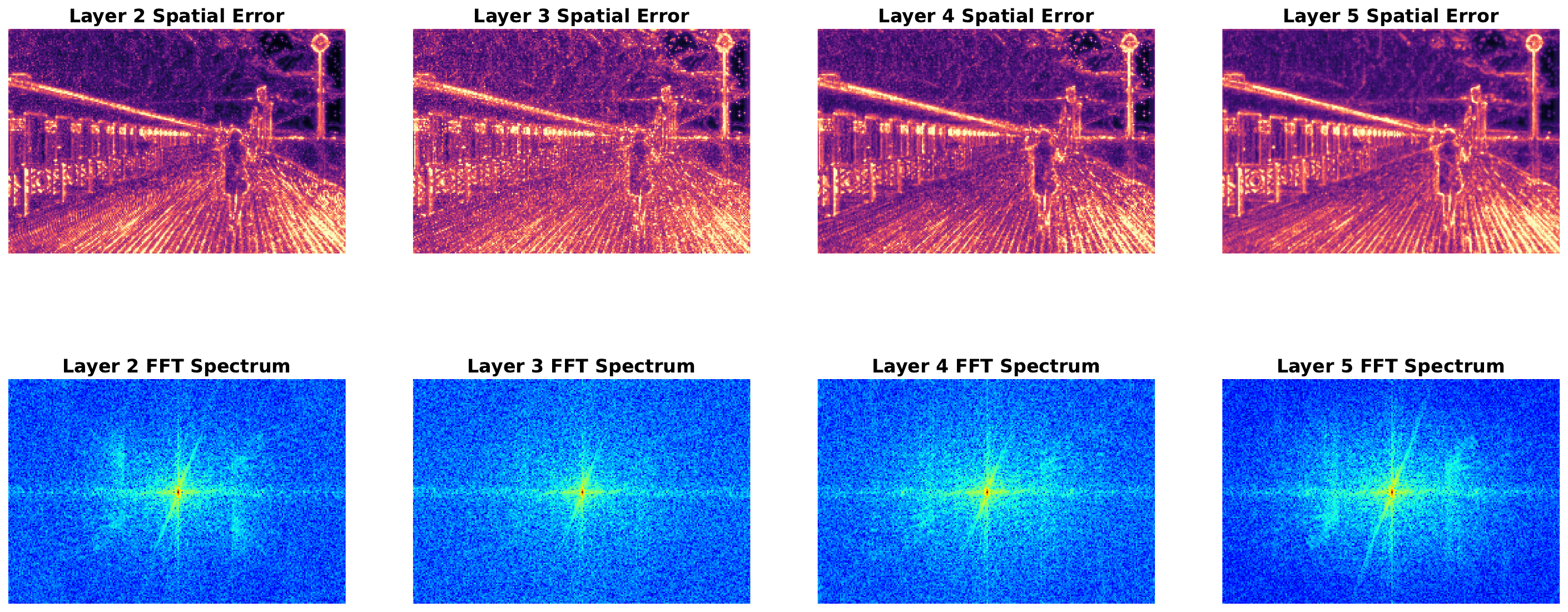}
        \caption{Urban100 img\_032}
        \label{fig:fft_2}
    \end{subfigure}

    \vspace{1em} 

    \begin{subfigure}{0.48\textwidth}
        \centering
        \includegraphics[width=\textwidth]{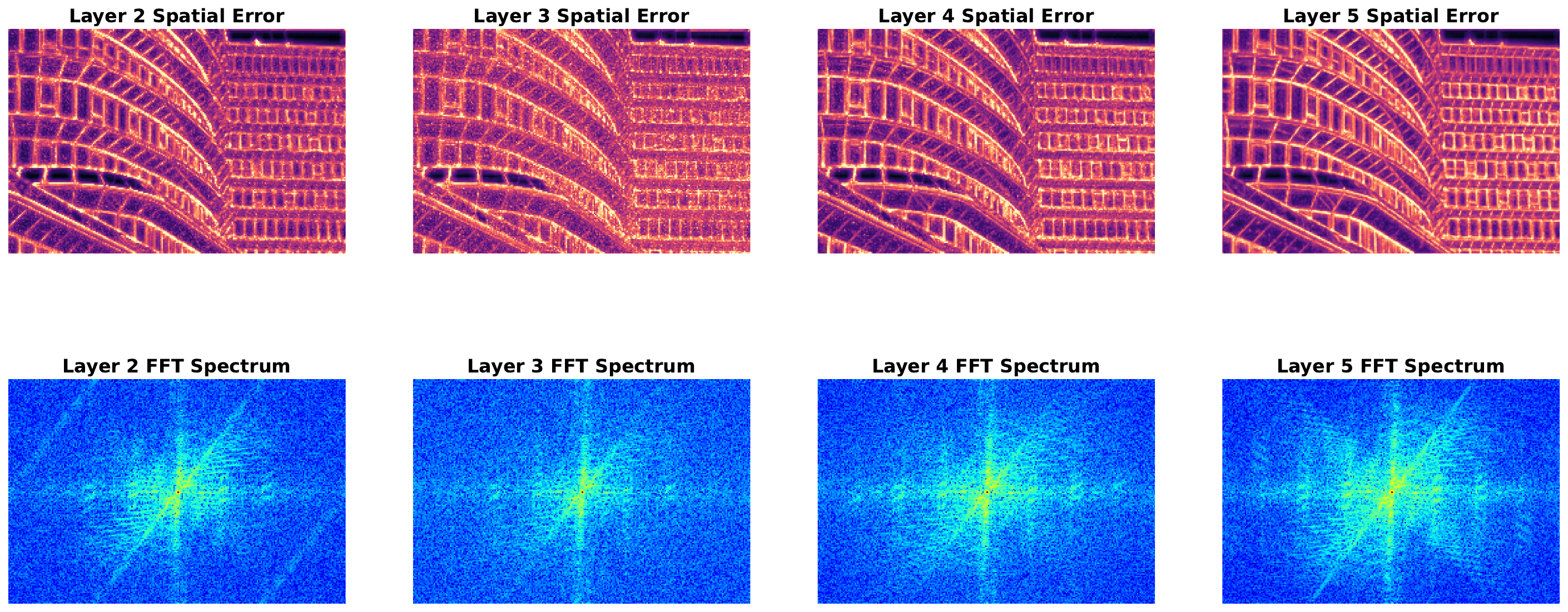}
        \caption{Urban100 img\_052}
        \label{fig:fft_3}
    \end{subfigure}
    \hfill
    \begin{subfigure}{0.48\textwidth}
        \centering
        \includegraphics[width=\textwidth]{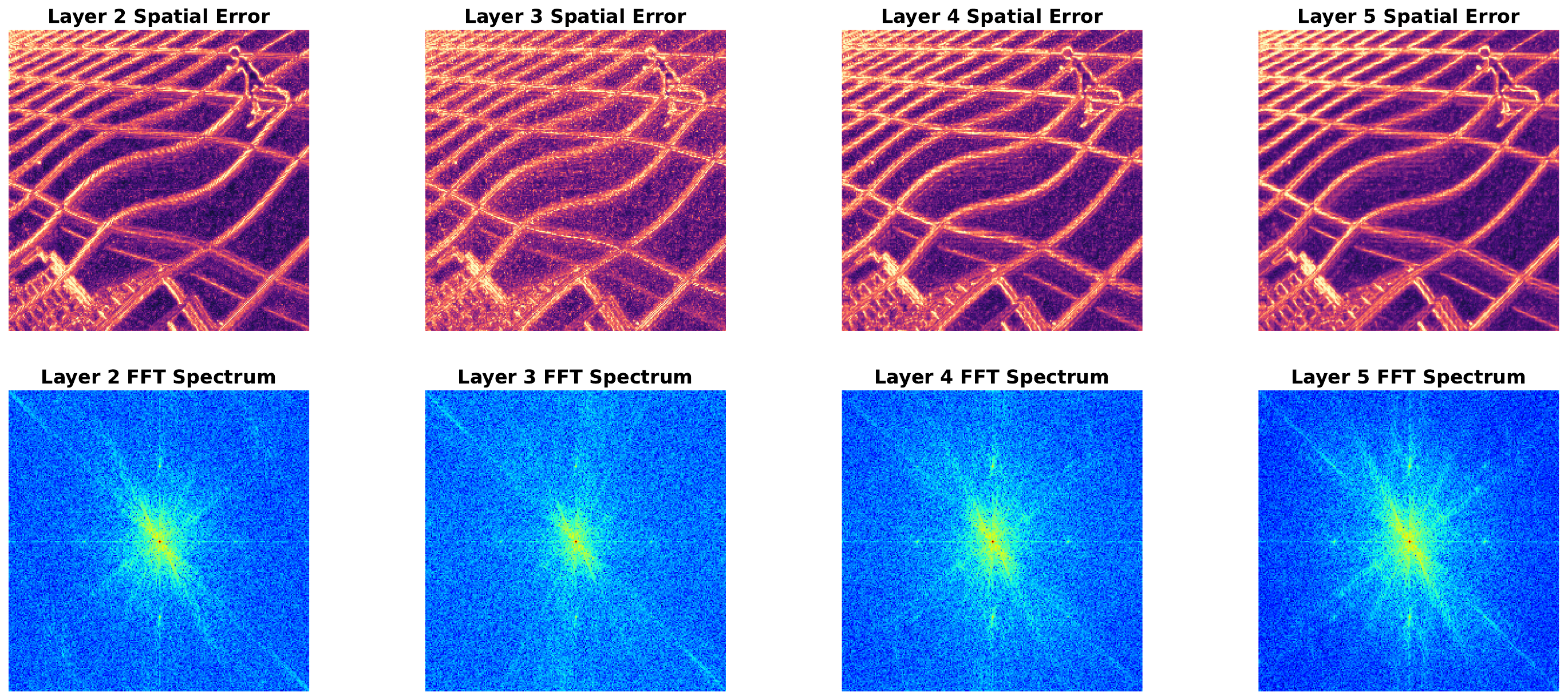}
        \caption{Urban100 img\_082}
        \label{fig:fft_4}
    \end{subfigure}

    \caption{
    Visualization of spatial error maps and their corresponding FFT spectra at layers 2, 3, 4, and 5 on the Urban100 dataset. The frequency-domain analysis highlights the model's systematic refinement of high-frequency information, illustrating the transition from suppressing redundant low-frequency priors to prioritizing high-fidelity edge restoration.}
    \label{fig:app_fft}
\end{figure}

In this section, we provide an in-depth analysis of the internal characteristics arising from the structural differences between the CUST-Base and CUST-Base+ models. While CUST-Base employs variable MEDA window sizes—periodically cycling through [12, 14, 16, 18]—CUST-Base+ utilizes a fixed window size of 18. We investigate how these configurations influence the internal window dynamics of the MEDA module in Figure~\ref{fig:app_win_sim}.

Figure~\ref{fig:app_win_sim} visualizes the intra-window patch similarity, representing the consistency of patches within each window across different datasets. The x-axis denotes the layer index, the y-axis represents the analyzed window indices, and the color intensity indicates the average cosine similarity between patches within a window. Our analysis reveals that CUST-Base exhibits higher similarity in the initial stages (Layers 0-1) compared to CUST-Base+. In contrast, CUST-Base+ shows a trend of increasing intra-window similarity toward the final layers, particularly in the Set5 and Set14 datasets. This suggests that CUST-Base achieves higher initial consistency through more localized and precise partitioning via variable window sizes. Conversely, while CUST-Base+ may experience lower initial consistency due to the grouping of heterogeneous patches within larger fixed windows, it achieves superior global feature alignment as the network deepens.

Furthermore, we perform a frequency-domain analysis using Fourier Transform, as illustrated in Figure~\ref{fig:app_fft}, to investigate the temporary decline in structural and content-level similarity at specific layers. As shown in Fig.~\ref{fig:app_win_sim}, a structural bottleneck occurs at Layer 2, where intra-window similarity drops across all datasets. This is immediately followed by a decline in content-level similarity at Layer 3, as evidenced in Fig.~\ref{fig:app_meda_align}. Frequency-domain analysis at Layer 3 reveals a temporary dispersion of the FFT spectrum. We interpret this as a phase where the model suppresses redundant low-frequency structural priors to prioritize high-frequency (HF) restoration. By Layer 4, the spectrum becomes strongly realigned along dominant edge directions, which is reflected in the recovery of intra-window similarity (Fig.~\ref{fig:app_win_sim}) and edge-wise similarity (Fig.~\ref{fig:app_meda_align}). Through this systematic information refinement, MEDA actively extracts features optimized for high-fidelity restoration.

\end{document}